\documentclass[lettersize,journal]{IEEEtran}
\usepackage{graphicx}
\usepackage{subcaption}
\usepackage{algorithm}
\usepackage{algorithmicx}
\usepackage{algpseudocode}
\usepackage{amssymb}
\usepackage{amsmath       }
\usepackage{cite}
\usepackage{setspace}
\usepackage{longtable}
\usepackage{diagbox}
\usepackage{multirow}
\usepackage{color}
\usepackage[switch]{lineno}
\usepackage{amsfonts}
\usepackage{amssymb}
\usepackage{amsthm}
\usepackage{mathrsfs}
\usepackage{indentfirst}
\usepackage{paralist}
\usepackage[bookmarks=true, hidelinks]{hyperref}
\usepackage{makecell}

\newcommand{\x}{\mathbf{x}}
\newcommand{\w}{\mathbf{w}}
\newcommand{\vb}{\mathbf{v}}
\newcommand{\rb}{\mathbf{r}}
\newcommand{\gb}{\mathbf{g}}
\def\mask{\texttt{MASK}}

\DeclareUnicodeCharacter{2061}{}
\begin{document}
\title{Attacking Important Pixels for Anchor-free Detectors }

\author{
Yunxu Xie$^{*}$,
Shu Hu$^{*}$,  
Xin Wang$^{\ddagger}$, \IEEEmembership{Senior Member,~IEEE}, \\
Quanyu Liao,
Bin Zhu,
Xi Wu$^{\ddagger}$,
Siwei Lyu \IEEEmembership{Fellow,~IEEE}
\thanks{Yunxu Xie, Quanyu Liao, and Xi Wu are with the Chengdu University of Information Technology, China. e-mail:(\{xieyunxu, xi.wu\}@imde.ac.cn)} 
\thanks{Shu Hu is with Carnegie Mellon University, USA. e-mail:shuhu@cmu.edu}
\thanks{Xin Wang and Siwei Lyu are with University at Buffalo, SUNY, USA. e-mail:(\{xwang264, siweilyu\}@buffalo.edu)}
\thanks{Bin Zhu is with Microsoft Research Asia. e-mail:(binzhu@microsoft.com)}
\thanks{$^{*}$ Equal contribution. $^{\ddagger}$ Corresponding authors.}
}


\maketitle

\begin{abstract}
Deep neural networks have been demonstrated to be vulnerable to adversarial attacks: subtle perturbation can completely change the prediction result. Existing adversarial attacks on object detection focus on attacking anchor-based detectors, which may not work well for anchor-free detectors. 
In this paper, we propose the first adversarial attack dedicated to anchor-free detectors. It is a category-wise attack that attacks important pixels of all instances of a category simultaneously.
Our attack manifests in two forms, sparse category-wise attack (SCA) and dense category-wise attack (DCA), that minimize the $L_0$ and $L_\infty$ norm-based perturbations, respectively. For DCA, we present three variants, DCA-G, DCA-L, and DCA-S, that select a global region, a local region, and a semantic region, respectively, to attack. Our experiments on large-scale benchmark datasets including PascalVOC, MS-COCO, and MS-COCO Keypoints indicate that our proposed methods achieve state-of-the-art attack performance and transferability on both object detection and human pose estimation tasks.
\end{abstract}

\IEEEpeerreviewmaketitle
\section{Introduction}
\IEEEPARstart{T}{he} development of deep neural networks (DNNs) \cite{xiang2022rmbench, pu2022learning, hu2022pseudoprop, guo2022robust, luo2022stochastic} enables researchers to achieve unprecedented high performance on various computer vision tasks. However, DNN models are vulnerable to adversarial examples: intentionally crafted subtle disturbance on a clean image makes a DNN model predict incorrectly~\cite{hu2021tkml,goodfellow2014FGSM,moosavi2016deepfool,kurakin2016pgd,dong2018boosting,modas2019sparsefool,we2018uea,xie2017dag,Sanchez2020RPFGSM,Shermin2021domainadaptation,hu2022rank, hu2021tkml}. 
As a typical type DNNs, convolutional neural networks (CNNs) achieve state-of-the-art (SOTA) performance on classification, objection detection, segmentation, and other computer vision tasks but also suffer from adversarial examples~\cite{szegedy2013intriguing,Qian2021CFR}. Studies on generating adversarial examples have attracted increasing attention because they help identify vulnerabilities of trained DNN models before they are launched in services. 


\begin{figure}[t]
\begin{center}
\includegraphics[width=0.47\textwidth]{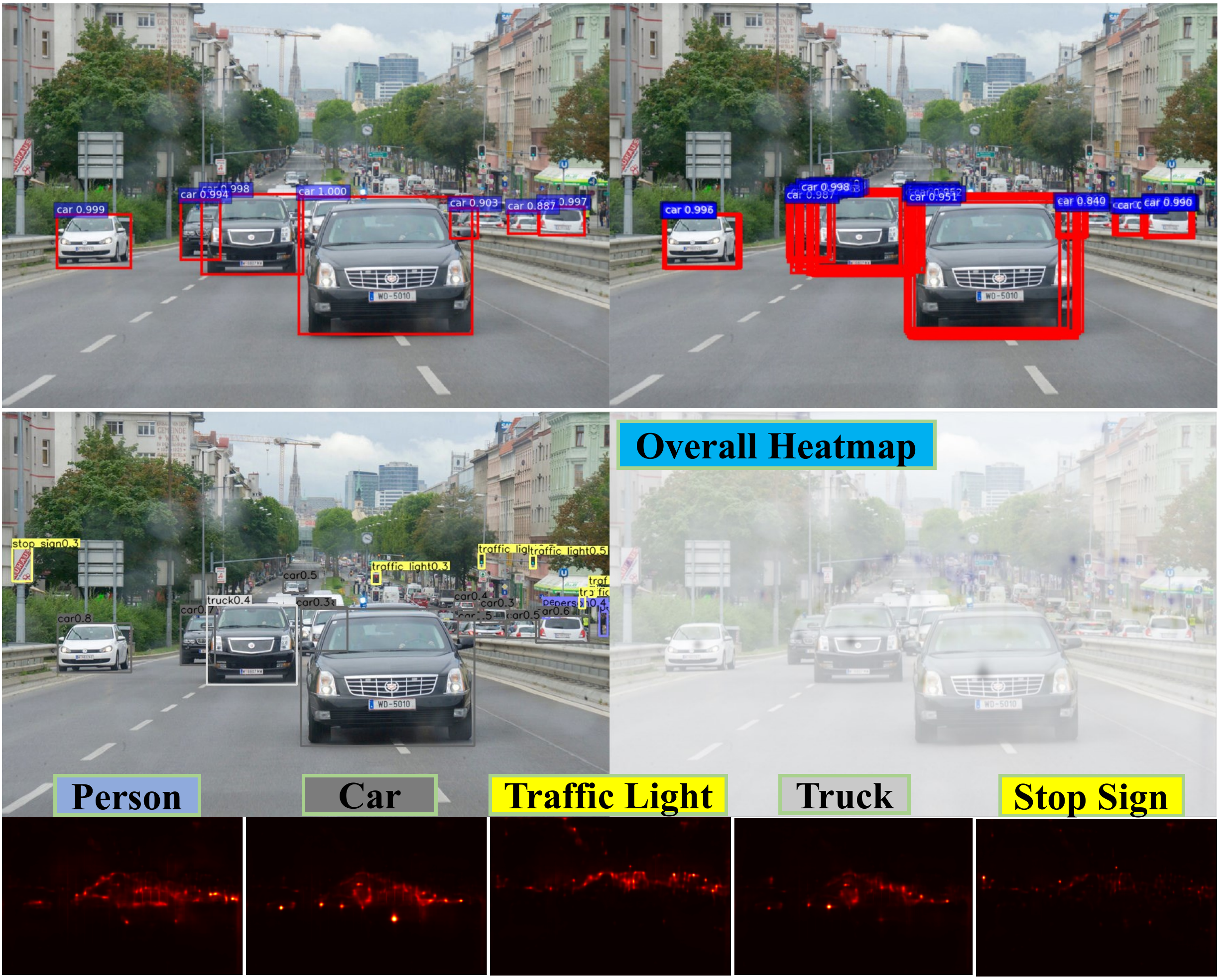}
\end{center}
\vspace*{-0.3cm}
\caption{ \small
\textbf{First row:} The detected results (left) and the proposals (right) of Faster-RCNN~\cite{ren2015fasterrcnn}.
\textbf{Second row:} The detected results (left) and the overall heatmap (right) of CenterNet~\cite{zhou2019centernet}. \textbf{Third row:} Selected target pixels (red) that will be attacked for each category by our methods.}
\label{fig_pix_select}
\vspace*{-0.5cm}
\end{figure}

Object detection is essential in many vision tasks like instance segmentation, pose estimation, and action recognition. Existing object detectors can be classified into two groups according to the way they locate objects: anchor-based~\cite{liu2016ssd,Li2018retinanetfocal,redmon2016ayolov2,ren2015fasterrcnn,he2017maskrcnn,2021dualattention} and anchor-free detectors~\cite{law2018cornernet,zhou2019bottom,zhou2019centernet,dong2020CentripetalNet,lin2022contrastive}. An anchor-based detector first determines many preset anchors on the image and then refines their coordinates and predicts their categories before outputting final detection results (see the first row in Fig.~\ref{fig_pix_select}). An anchor-free detector finds objects without using preset anchors: it detects keypoints of objects and then bounds their spatial extent (see Fig.~\ref{fig_pix_select} second row). 
There are many works on adversarial attacks against anchor-based detectors, such as DAG~\cite{xie2017dag} and UEA~\cite{we2018uea}. 
However, to the best of our knowledge, there is no published work except our published conference papers~\cite{liao2021transferable, liao2020fast} on investigating vulnerabilities of anchor-free detectors.

Attacking an anchor-free detector is very different from attacking an anchor-based detector. Adversarial attacks on anchor-based detectors work on selected top proposals from a set of anchors of objects, while anchor-free object detectors return only objects' keypoints via the heatmap mechanism (see the second row in Fig.~\ref{fig_pix_select}). These keypoints are used to generate corresponding bounding boxes. This detection procedure is completely different from anchor-based detectors, making anchor-based adversarial attacks unable to directly adapt to attack anchor-free detectors. In addition, many attack methods for anchor-based detectors such as DAG and UEA suffer from the following shortcomings. First, their generated adversarial examples have a poor transferability, \emph{e.g.}, adversarial examples generated by DAG on Faster-RCNN 
can hardly be transferred to attack other object detectors. Second, some of them attack only one proposal at a time, which is extremely costly in computation. Thus, it is desired to investigate new efficient and effective attack schemes for anchor-free detectors.

In this work, 
we propose a novel untargeted adversarial attack, called \emph{Category-wise Attack}, to attack anchor-free detectors. Our proposed attack focuses on categories and can attack all objects from the same category simultaneously by attacking a set of important target pixels (see the third row in Fig.~\ref{fig_pix_select}). 
The important target pixel set includes detected pixels that are highly informative (higher-level information of objects)
as well as undetected pixels that have a high probability to be detected.

Our category-wise attack is formulated as a general framework that minimizes $L_p$ norm-based perturbations, where $p\in\{0,\infty\}$, to flexibly generate 
sparse and dense perturbations, called 
\textit{sparse category-wise attack} (SCA) and \textit{dense category-wise attack} (DCA), respectively. Moreover, in DCA, we explore three attack strategies based on different attack regions: DCA on the global region (DCA-G) that attacks the whole region of an image, DCA on the local region (DCA-L) that attacks only important regions around  objects, and DCA on the semantic region (DCA-S) that attacks semantic-rich regions of objects. Attacking only specific important regions can effectively reduce the number of pixels disturbed while retaining high attack performance.

We demonstrate the effectiveness of our
methods in attacking anchor-free detector CenterNet~\cite{zhou2019centernet} with different backbones in both object detection and human pose estimation tasks using large scale benchmark datasets: PascalVOC~\cite{everingham2015pascal}, MS-COCO~\cite{lin2014microsoftcoco}, and COCO-keypoints~\cite{lin2014microsoftcoco}.
Our experimental results show that our attack methods outperform existing SOTA methods with high attack performance and low visibility of perturbations.

The main contributions of
our work can be summarized as follows:
\begin{enumerate}
\item We present the first algorithms of untargeted adversarial attacks on anchor-free detectors. They attack all objects in the same category simultaneously instead of only one object at a time, which avoids perturbation over-fitting on one object and increases the transferability of generated adversarial examples.

\item Our category-wise attack is designed to attack important pixels in images. On one hand, it can generate sparse adversarial perturbations to increase imperceptibility of generated adversarial examples. On the other hand, it can generate dense adversarial perturbations to improve attacking effectiveness.   

\item Our method generates more transferable adversarial examples than existing attacks in both object detection and human pose estimation tasks. Our experiments on large-scale benchmark datasets indicate that it achieves the SOTA performance for both white-box and black-box attacks.

\end{enumerate}

This paper extends our published conference papers \cite{liao2021transferable} (ICME 2021) and \cite{liao2020fast} (IJCNN 2020) substantially in the following aspects: 
1) We provide a general DCA algorithm that includes three attacking approaches based on the targeted attacking region. In particular, we propose a new DCA method called DCA-S that restricts perturbations in the semantic region, which is extracted according to informative gradients. DCA-S can generate more semantically meaningful perturbations. 
2) We show the applicability of our SCA and DCA attack methods in a practical human pose estimation task on the MS-COCO Keypoints dataset and demonstrate that they outperform existing attack methods.
3) We add more experiments to verify the effectiveness of our proposed attacking methods, especially for DCA-S. We also add more studies for analyzing the sensitivity of hyperparameters.     

The rest of the paper is organized as follows. In Section~\ref{sec: related_work}, we discuss the difference between anchor-free and anchor-based detectors and summarize existing adversarial attacks that are related to this work. In Section~\ref{sec:problem_formulation}, we formulate our attacking problem in the category-wise attack setting for anchor-free detectors. In Section~\ref{sec:methodology}, we provide a detailed description of our proposed category-wise attack in terms of sparse and dense settings. In Section~\ref{sec:experiments}, we conduct experiments to demonstrate the efficiency and effectiveness of our proposed algorithms for attacking anchor-free detectors in both object detection and human pose estimation tasks. Section~\ref{sec:conclusion} concludes the paper with discussions.

\section{Related work}
\label{sec: related_work}

\subsection{Anchor-based and Anchor-free Detectors}
\noindent

A great progress has been made in object detection in recent years. With the development of deep convolutional neural networks, many object detection approaches have been proposed.
One of the most popular groups of object detection methods is the RCNN~\cite{Girshick2013rcnn} family, such as Faster-RCNN~\cite{ren2015fasterrcnn}. The first process of RCNN's pipeline is to generate a large number of proposals based on anchors. Then a different classifier is used to classify the proposals. At last, a post-processing algorithm, such as non-maximum
suppression (NMS)~\cite{neubeck2006efficient}, is used to reduce redundancy proposals. Other typical anchor-based object detectors include YOLOv2~\cite{redmon2016ayolov2} and SSD~\cite{liu2016ssd}. They need to place a set of rectangles with pre-defined sizes during the training and then put them in some desired positions. Anchor-base detectors have high detection accuracy but also have the following shortcomings: anchor boxes should be manually defined before the training and these anchor boxes may not coincide and be consistent with ground-truth boxes.

To solve these shortcomings, anchor-free object detectors have been proposed, such as CornerNet~\cite{law2018cornernet} and CenterNet~\cite{zhou2019centernet}. These object detectors detect objects by detecting their keypoints. Specifically, CornerNet detects an object by detecting the two corners of the object, and CenterNet detects objects by finding their center points. No anchor is used in both detectors. These two detectors not only are faster and simpler to train than anchor-based detectors but also achieve SOTA detection performance. Both CornerNet and CenterNet can use multiple convolutional neural networks as the backbone network to extract semantic features of an input image. Keypoints of objects are then located through these features. Normally, a keypoint includes the size and category (or class) information of its object. At last, some post-processing algorithm is used to remove redundancy keypoints. With its good keypoint estimation, CenterNet works not only for object detection but also for human pose estimation.

\subsection{Adversarial Attacks on Object Detection}
\noindent

Goodfellow et al.~\cite{goodfellow2014FGSM} first showed the adversarial example problem of deep neural networks. An adversarial example is an original sample perturbed with deliberately crafted perturbation, typically imperceptible, that makes the DNN model predict incorrectly.
Adversarial attacks can be classified into two groups: targeted adversarial attacks and untargeted adversarial attacks. A former attack aims to make the DNN model to mispredict to a specific label while a latter attack aims to make the DNN model to mispredict to anything different from the original label (including failure to detect objects for object detection). Our proposed attack is an untargeted adversarial attack.

Most adversarial attacks focus on minimizing the $\mathop{L_p}$ ($p \in (0, 1, 2, \infty)$) norm of the adversarial perturbation, aiming to make the perturbation imperceptible while maintaining a high attack success rate. Typical adversarial attacks include Fast Gradient Sign Method (FGSM)~\cite{goodfellow2014FGSM}, Project Gradient Descent (PGD)~\cite{kurakin2016pgd}, 
and DeepFool~\cite{moosavi2016deepfool}. Specifically, 
FGSM computes the gradient of the loss with respect to the input image to generate the adversarial perturbation as follows:
\begin{equation*}
    \mathop{\overline{\x} = \x + \epsilon\cdot sign(\nabla_{\x}J(f(\x),y))},
    \label{ep1}
\end{equation*}
where $f$ is the classifier, $J$ is the loss function, $\x$ is the original input image, $y$ is the ground-truth label of the input image, $\overline{\x}$ is the perturbed image, and  $\epsilon$ is the amplitude of the perturbation. 
PGD applies FGSM iteratively with a smaller amplitude of the perturbation $\alpha$:
\begin{equation*}
    \x_{t} = \Pi_{\mathcal{B}_\epsilon(\x)}\big(\x_{t-1} + \alpha\cdot sign(\nabla_{\x}J(f(\x_{t-1}),y)\big),
\end{equation*}
where $\x_0=\x$ and $\Pi_{\mathcal{B}_\epsilon(\x)}(\cdot)$ is the projection function that projects the perturbed image back into the $\epsilon$-ball centered at the original input image $\x$ if necessary. 
DeepFool uses a generated hyperplane to approximate the decision boundary and computes the lowest Euclidean distance between the input image and the hyperplane iteratively. It then uses the distance to generate the adversarial perturbation. 


The above attack methods mainly attack classifiers. Adversarial attacks on object detectors have also been proposed, such as DAG~\cite{xie2017dag} and UEA~\cite{we2018uea}. They all focus on attacking anchor-base object detectors such as Faster-RCNN and SSD. 
The main shortcomings of 
these adversarial attacks are slow and not robust for transferring attacks. Furthermore, they are designed for attacking anchor-based detectors and don't work well for attacking anchor-free object detectors.
To the best of our knowledge, there are no published adversarial attack on anchor-free detectors except our two conference papers~\cite{liao2021transferable, liao2020fast} that this paper extends.

\begin{figure}[t]
    \begin{center}
    \includegraphics[width=0.47\textwidth]{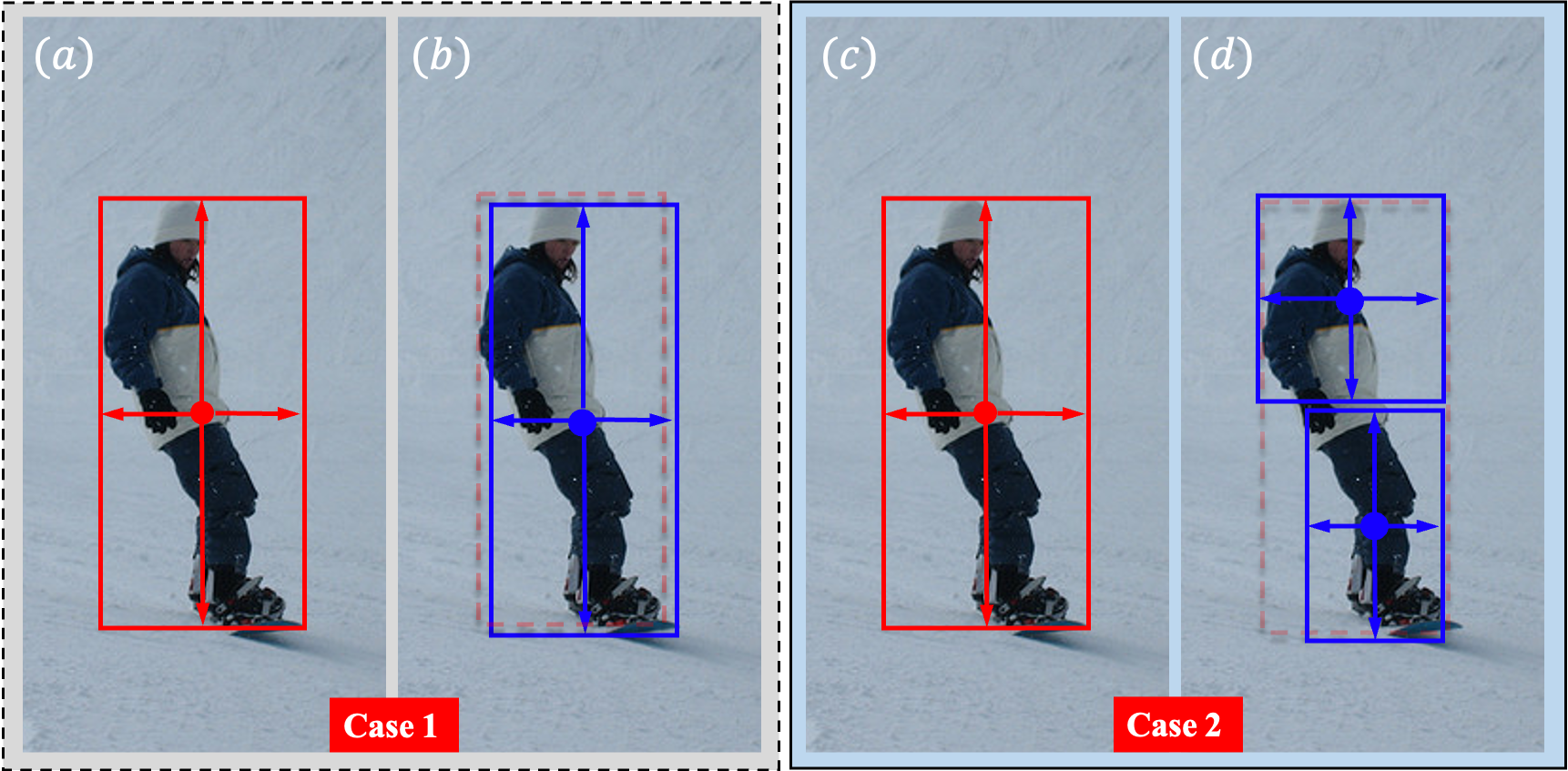}
    \end{center}
    \vspace*{-0.2cm}
    \caption{ \small
    Detection results before and after attacking only detected pixels.
    \textbf{Red} boxes denote originally detected keypoints and bounding boxes before the attack.
    \textbf{Blue} boxes denote newly detected keypoints after the attack.
    \textbf{(a) \& (c):} a detected object and a detected keypoint at the center of the person before the attack in cases 1 and 2.
    \textbf{(b) \& (d):} detection results after attacking only detected pixels. In case 1, after attacking all detected pixels, a neighboring pixel of the previously detected keypoint is detected as the correct category. In case 2, the centers of the top half and the bottom half of the person appear as newly detected keypoints still detected as a person. In both cases, mAP is barely reduced.}
    \label{select_pixel}
    \vspace*{-0.5cm}
\end{figure}

\section{Problem Formulation}\label{sec:problem_formulation}

In this section, we define our optimization problem of attacking anchor-free detectors based on category-wise attacks.

Suppose there exist $k$ object categories, $\mathcal{C}:=\{C_1, C_2, ..., C_k\}$, with detected object instances. Let $[k]:=\{1,...,k\}$. We use $S_{i}$ to denote the target pixel set of category $C_{i}$ whose detected object instances will be attacked. There are $k$ target pixel sets,  $\mathcal{S}:=\{S_1, S_2, ..., S_k\}$. 
The \textbf{category-wise attack} for anchor-free detectors is formulated as the following constrained optimization problem:
\begin{equation}
   \begin{aligned}
       \mathop{\text{minimize}} \limits_{\rb} \quad & \Vert{\rb} \Vert_{p}, \\
       \text{s.t.} ~ \quad & {\arg\max}_j f_j(\x+\rb, s) \not = C_{i}, s\in S_{i}, \forall i\in [k],  \\
   \end{aligned}
   \label{categorywiseoptimization}
\end{equation}
where $\rb$ is an adversarial perturbation, $\Vert{\cdot}\Vert_{p}$ is the $L_p$ norm, 
$\x$ is a clean input image, $\x+\rb$ is an adversarial example, $f(\x+\rb, s)$ is the classification prediction score vector for pixel $s$ and $f_j(\x+\rb, s)$ is its $j$-th value, where $j\in[k]$, and 
${\arg\max}_jf_j(\x+\rb, s)$ denotes the predicted object category on a target
pixel $s \in S_{i}$ of adversarial example $\x+\rb$. While $p$ in the the $L_p$ norm can take different values, we focus on $p \in \{0, \infty\}$ in this paper.

In solving Eq.~\ref{categorywiseoptimization}, it is natural to use all \emph{detected pixels} of category $C_{i}$ as target pixel set $S_{i}$.  The {detected pixels} are selected from the heatmap of category $C_{i}$ generated by an anchor-free detector such as CenterNet \cite{zhou2019centernet} with their probability scores higher than the detector's preset visual threshold and being detected as right objects. Unfortunately, it does not work: after attacking all detected pixels to predict into wrong categories, we expect that the detector should not detect any correct object, but our experiments with CenterNet turn out that it still can.

\begin{figure*}[t]
    \begin{center}
    \includegraphics[width=0.97\textwidth]{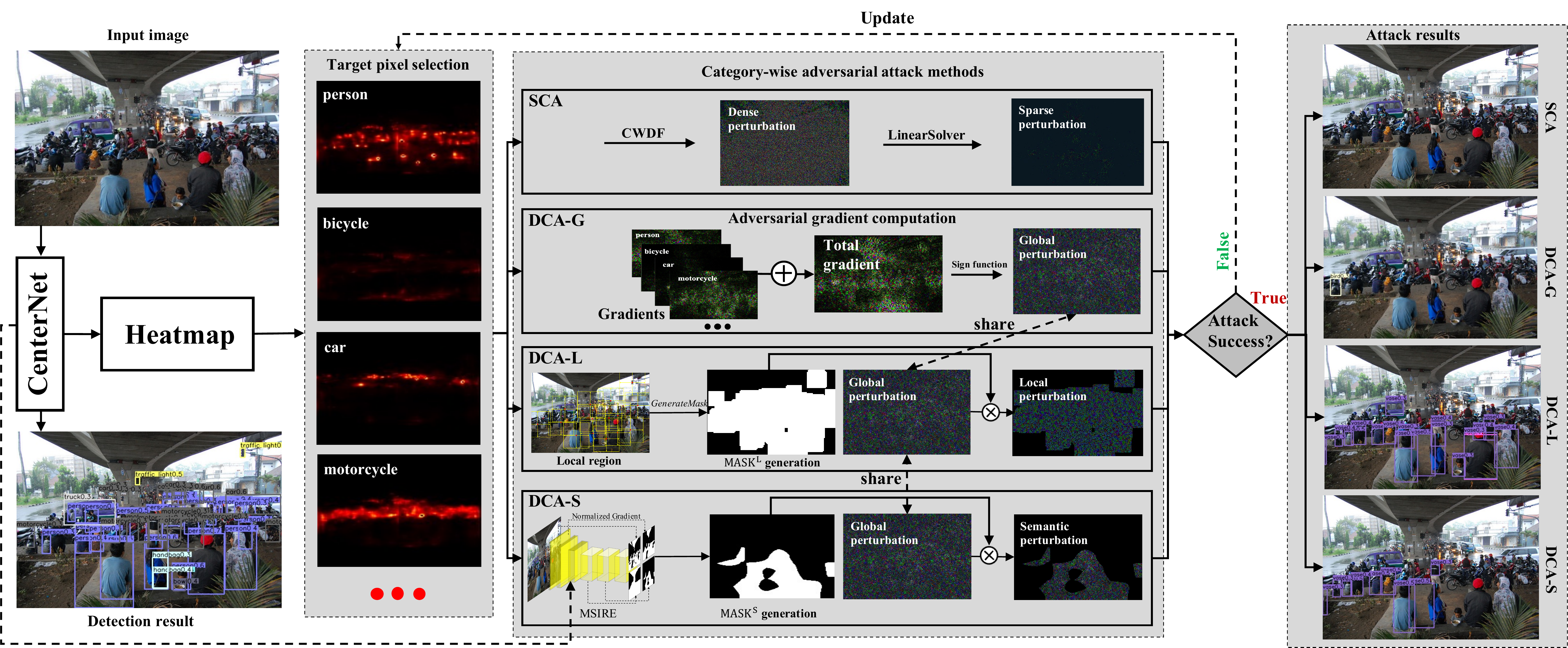}
    \end{center}
    \vspace*{-0.2cm}
    \caption{\small Our general attack framework overview. Firstly, we extract the heatmap for each object category. Target pixels are selected based on heatmap and attacking threshold $\mathcal{T}$. Secondly, our proposed category-wise adversarial attack methods including sparse category-wise attack (SCA) and dense category-wise Attacks (DCA-G, DCA-L, and DCA-S) are used to attack target pixels from each category. More details about these methods can be found in the main paper. Finally, target pixel sets will be updated and the perturbation will be delicately modified if the attack is not complete. Otherwise, the attack results will output.
    }
    \label{fig4_overview}
    \vspace*{-0.5cm}
\end{figure*}

Further investigation enables us to explain the unexpected result as follows: 
\begin{enumerate}
    \item Unattacked neighboring background pixels of the heatmap can become detected pixels of the correct category. Since their detected box is close to the old detected object, CenterNet can still detect the object even though all the previously detected pixels are attacked into wrong categories. An example is shown in Fig.~\ref{select_pixel} (a) and (b).
    \item CenterNet regards center pixels of an object as keypoints. After attacking detected pixels located around the center of an object, newly detected pixels may appear in other positions of the object, making the detector still be able to detect multiple local parts of the correct object with barely reduced mAP. An example is shown in Fig.~\ref{select_pixel} (c) and (d).
\end{enumerate}

Pixels that can produce one of the above two changes are referred to as \emph{runner-up pixels}. We find that almost all \emph{runner-up pixels} have a common characteristic: their probability scores are only a little below the visual threshold used in the object detector. Based on this characteristic, our category-wise attack methods set an attacking threshold, $\mathcal{T}$, lower than the visual threshold, and select all the pixels from the heatmap whose probability score is above $\mathcal{T}$ into $S_i$. Therefore, $S_i$ includes all detected and \emph{runner-up pixels}. In this way, we can improve the  transferability of generated adversarial examples.

\section{Methodology}\label{sec:methodology} 
In this section, we provide a detailed description of our category-wise attack, which is a sparse category-wise attack (SCA) if $L_0$ is used and a dense category-wise attack (DCA) if $L_{\infty}$ is used. In DCA, we explore and design three attack strategies called DCA on the global region (DCA-G), DCA on the local region (DCA-L), and DCA on the semantic region (DCA-S). An overview of the proposed category-wise attack (CA) is shown in Fig.~\ref{fig4_overview}. In the following description of our methods, we focus on untargeted adversarial attack tasks. If the task is a targeted adversarial attack, our methods can be described in a similar manner.

\subsection{Sparse Category-wise Attack (SCA)}
The goal of the sparse category-wise attack is to fool the detector while perturbing a minimum number of pixels in the input image. It is equivalent to setting $p=0$ in 
Eq.~\ref{categorywiseoptimization}. Unfortunately, this is an NP-hard problem. To solve this problem,  SparseFool~\cite{modas2019sparsefool} relaxes this NP-hard problem by iteratively approximating the classifier as a local linear function in generating sparse adversarial perturbation for image classification. Motivated by the success of SparseFool on image classification, we propose Sparse Category-wise Attack (SCA) to generate sparse perturbations for anchor-free object detectors. It is an iterative process. In each iteration, one target pixel set is selected to attack. 

More specifically, given an input image $\x$ and current category-wise target pixel sets $\mathcal{S}$, the pixel set $S_{h}$ that has the highest probability score from $\mathcal{S}$ is selected, where $h\in[k]$. We can formulate $S_{h}$ as 
\begin{equation}
   \begin{aligned}
       S_{h} = \arg\max_{S_i}\Bigg\{\sum_{s\in S_i}f_{i}(\x,s)\Bigg|S_i\in \mathcal{S}, i\in [k]\Bigg\}.
   \end{aligned}
\label{eq:target_set}
\end{equation}

\renewcommand{\algorithmicrequire}{\textbf{Input:}}
\renewcommand{\algorithmicensure}{\textbf{Output:}}
\begin{algorithm}[t]
    \setstretch{1.3}
    \caption{Category-Wise DeepFool (CWDF)}
    \label{df}
    \begin{algorithmic}

        \Require
        image $\x$, target pixel set $S_{h}$, $\mathcal{T}$
        \Ensure
        dense adversarial example $\x^B$

        \State{\textbf{Initialize}: $\x_1 \leftarrow \x, p \leftarrow 1$}

        \While{$S_{h} \not= \varnothing$}
        
            \For{$j \not= h$}

                \State$\vb_j \leftarrow \nabla\sum_{s \in S_{h}}{f_{j}(\x_p, s)}-\nabla\sum_{s \in S_{h}}{f_{h}(\x_p, s)}$
                
                \State$\text{score}_j \leftarrow \sum_{s \in S_{h}}f_j(\x_p, s)$
            
            \EndFor
            
            \State$o \leftarrow \text{argmin}_{j\not=h}\ \frac{\vert \text{score}_j\vert}{\Vert \vb_j\Vert_2}$,
            $\x_{p+1} \leftarrow \x_p + \frac{\vert \text{score}_o\vert}{\Vert \vb_o\Vert_2^2}\vb_o$
            \State$S_{h} =$ \text{RemovePixels}($\x_p, \x_{p+1}, S_{h}, \mathcal{T}$) ~{\footnotesize/*Refer to Eq.~ \ref{eq:update_target}*/}
            \State$p \leftarrow p + 1$
        \EndWhile
 
        \State{\textbf{return} $\x^B \leftarrow \x_{p}$ }
        
    \end{algorithmic}
\label{Alg:CWDF}
\end{algorithm}

Then we apply the \textit{Category-Wise DeepFool} (CWDF) method, whose algorithm is described in Algorithm~\ref{Alg:CWDF}, to generate a dense adversarial example $\x^B$ for $\x$ by computing perturbation on $S_{h}$ as follows,
\begin{equation}
   \begin{aligned}
       \x^B = \text{CWDF}(\x, S_{h}, \mathcal{T}).
   \end{aligned}
\label{eq:CWDF}
\end{equation}
CWDF is adapted from DeepFool~\cite{moosavi2016deepfool} to enable it to attack all pixels from target pixel set $S_h$ simultaneously. After the generation, we remove successfully attacked pixels from $S_h$.

\begin{figure}[t]
   \begin{center}
   \includegraphics[width=.9\linewidth]{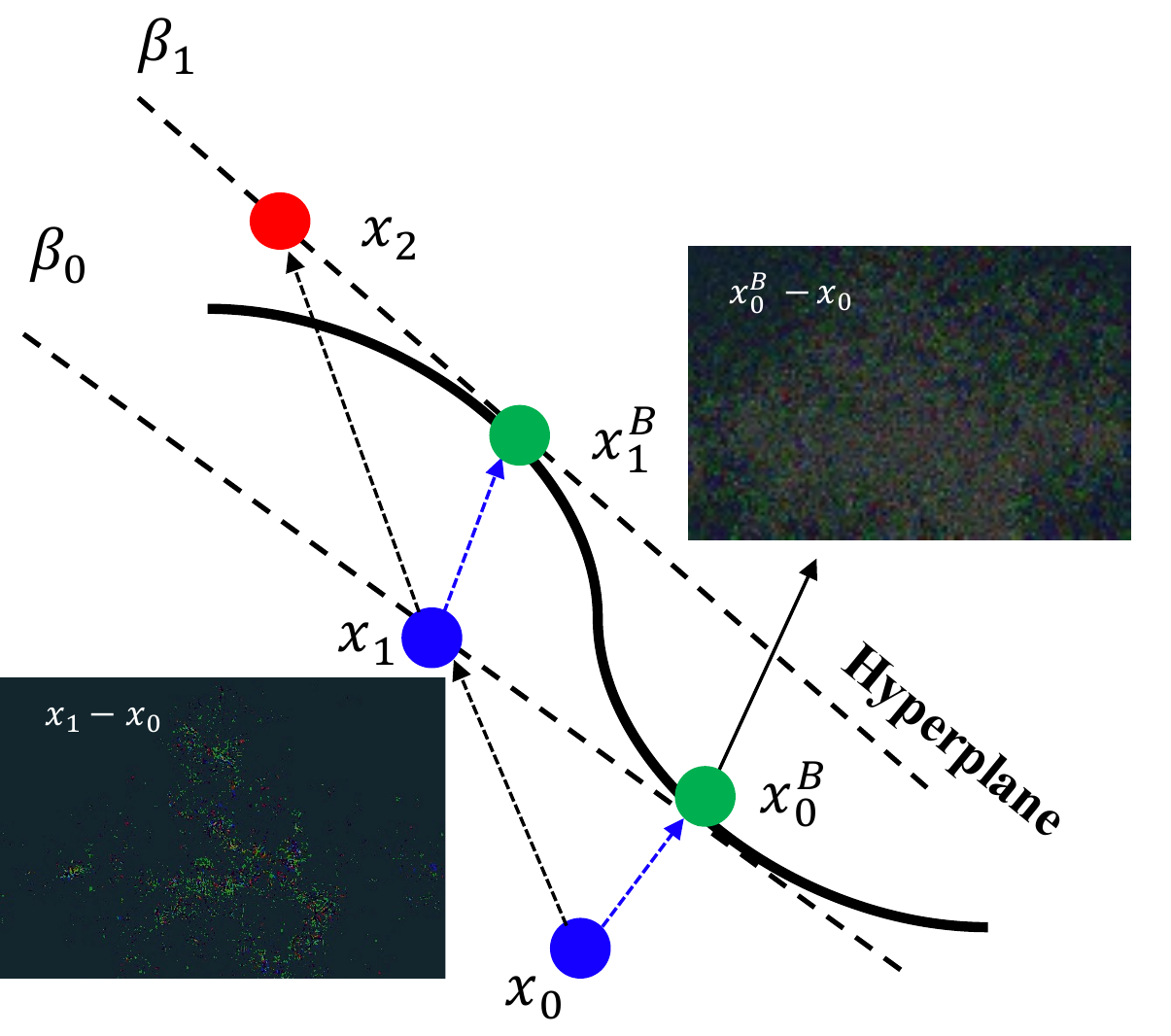}
   \end{center}
   \vspace*{-0.3cm}   
   \caption{\small Illustration of the \textit{ApproxBoundary} and the \textit{LinearSolver} of SCA. The black solid line denotes the real decision boundary of the object detector. Blue points denote adversarial examples that have not attacked all objects successfully. Red point denotes adversarial examples that have already attacked all objects successfully. This figure illustrates two iterations of the attack, $\x_0 \rightarrow \x_1$ and $\x_1 \rightarrow \x_2$. Take $\x_0 \rightarrow \x_1$ as an example, SCA first generates a dense adversarial example $\x_0^B$ (green point) by \textit{CWDF} and approximated linear decision boundary $\beta_0$ (black dash line). Then it uses \textit{LinearSolver} to add a sparse perturbation to support $\x_0$ to approximate decision boundary $\beta_0$ by satisfying $\beta_0 = \{\x': \w^T(\x' - \x_0^B)= 0 \}$ until a valid sparse adversarial example $\x_1$ is obtained. The right image represents the perturbation after applying CWDF. The left image represents the perturbation after applying the \textit{ApproxBoundary} and the \textit{LinearSolver}. Comparing these two images, it is clear that the perturbation becomes sparse. 
   }
   \label{fig_sca_car}
  \vspace*{-.5cm}
\end{figure}

Next, SCA uses the \textit{ApproxBoundary} from~\cite{modas2019sparsefool} to approximate the decision boundary locally with a hyperplane $\beta$ passing through $\x^B$: $\beta \overset{\triangle}{=} \{\x':\w^T(\x' - \x^B) = 0 \}$,
where $\w$ is the normal vector of hyperplane $\beta$:
\begin{equation}\small
   \begin{aligned}
       \w\!\!&=\text{ApproxBoundary}(\x^{B}, \x, S_{h})\\
       &=\!\! \frac{\nabla\sum_{s\in S_{h}}  [{f_{\text{argmax}_j{f_j(\x^B, s)}}(\x^B, s)} 
       - {f_{\text{argmax}_j{f_j(\x, s)}}(\x^B, s)}]}{\|\nabla\sum_{s\in S_{h}}  [{f_{\text{argmax}_j{f_j(\x^B, s)}}(\x^B, s)} 
       - {f_{\text{argmax}_j{f_j(\x, s)}}(\x^B, s)}]\|_2}.
\end{aligned}
\label{eq:approxboundary}
\end{equation}
Sparse adversarial example $\x^{adv}$ can then be computed via the \textit{LinearSolver} process from~\cite{modas2019sparsefool}:
\begin{equation}
   \begin{aligned}
       \x^{adv}=\text{LinearSolver}(\x, \w, \x^{B}).
\end{aligned}
\label{eq:linearsolver}
\end{equation}
Specifically, in each iteration, we project $\x$ towards one single coordinate of $\w$. If the projection in a specific direction has no solutions, then we will ignore that direction in the next iteration because it cannot provide a significant contribution to the final perturbed image. More details can be found in~\cite{modas2019sparsefool}. 
For completeness, we include the algorithm of \textit{LinearSolver} in Appendix \ref{Sec:LinearSolver}.
The process of generating perturbation through the \textit{ApproxBoundary} and the \textit{LinearSolver} of SCA is illustrated in Fig.~\ref{fig_sca_car}.                     

After attacking $S_{h}$ through the above operations, SCA updates $S_{h}$ by removing pixels that are no longer detected. This process is called \textit{RemovePixels}. Specifically, taking $\x$, $\x^{adv}$, and $S_{h}$ as input, \textit{RemovePixels} first generates a new heatmap for perturbed image $\x^{adv}$ with the detector. Then, it checks whether the probability score of each pixel in $S_{h}$ is still higher than $\mathcal{T}$ on the new heatmap. Pixels whose probability score is lower than $\mathcal{T}$ are removed from $S_{h}$, while the remaining pixels are retained in $S_{h}$. Target pixel set $S_{h}$ is thus updated. We can formulate the \textit{RemovePixels} procedure as follows,
\begin{equation}
   \begin{aligned}
       S_{h} &= \text{RemovePixels}(\x, \x^{adv}, S_{h}, \mathcal{T})\\
       &= \Big\{s\Big|\arg\max_jf_j(\x^{adv}, s) \!\!=\!\! \arg\max_jf_j(\x, s),\\
       &\ \ \ \ \ \ \ \ \ \ f_{\arg\max_jf_j(\x^{adv}, s)}(\x^{adv}, s)>\mathcal{T}, s\in S_{h}\Big\}.
   \end{aligned}
\label{eq:update_target}
\end{equation}
We iteratively execute the above procedures (i.e., Eqs.~\ref{eq:target_set}, \ref{eq:CWDF}, \ref{eq:approxboundary}, \ref{eq:linearsolver}, and \ref{eq:update_target}) until $\mathcal{S}\in \varnothing$, i.e., there is no correct object 
can be detected after the attack. The attack for all objects of $\x$ is thus successful, and we output the final generated adversarial example. 

\renewcommand{\algorithmicrequire}{\textbf{Input:}}
\renewcommand{\algorithmicensure}{\textbf{Output:}}
\begin{algorithm}[t]
    \setstretch{1.3}
    \caption{Sparse Category-wise Attack (SCA)}
    \label{sca}
    \begin{algorithmic}

        \Require
        image $\x$, $\mathcal{S}$, $max\_iter\_outer$, $max\_iter\_inner$
        \Ensure
        adversarial example $\x^*$, perturbation $\rb$

        \State{\textbf{Initialize:} $\x_1 \leftarrow \x, p \leftarrow 1$}

        \While{$\mathcal{S} \not\in \varnothing$ and $p\leq max\_iter\_outer$}
        
            \State Compute $S_h$ with Eq. (\ref{eq:target_set}) and $\x_p$, $q \leftarrow 0$, $\x_{p,q}\leftarrow \x_{p}$;

            \While{$q \leq max\_iter\_inner$  and $S_{h} \notin \varnothing $}

                \State$\x_{p,q}^B = \text{CWDF}(\x_{p,q}, S_h, \mathcal{T})$~{\footnotesize /*See Alg.~\ref{Alg:CWDF}*/}
                \State$\w = \text{ApproxBoundary}(\x_{p,q}^{B}, \x_{p,q}, S_{h})$~{\footnotesize/*See Eq.~\ref{eq:approxboundary}*/}
                \State$\x_{p,q}^{adv} = \text{LinearSolver}(\x_{p,q}, \w, \x_{p,q}^{B})$ ~{\footnotesize/*See Eq.~\ref{eq:linearsolver}*/}
                \State$S_{h} = \text{RemovePixels}(\x_{p,q}, \x^{adv}_{p,q}, S_{h}, \mathcal{T})$ ~{\footnotesize/*See Eq.~\ref{eq:update_target}*/}

                \State$q = q + 1$

            \EndWhile

            \State$\x_{p+1} \leftarrow \x_{p,q-1}^{adv}$, $p = p + 1$

        \EndWhile
 
        \State{\textbf{return} $\x^*=\x_p$, $\rb = \x_p - \x$ }
        
    \end{algorithmic}
    \label{algorithm_sca}
\end{algorithm}

The SCA algorithm is summarized in Algorithm \ref{algorithm_sca}. In an iteration, if SCA fails to attack any pixels of $S_{h}$ in the inner loop, SCA will attack the same $S_{h}$ in the next iteration. During this process, SCA keeps accumulating perturbations on these pixels, with the probability score of each pixel in $S_{h}$ keeping reducing, until the probability score of every pixel in $S_{h}$ is lower than $\mathcal{T}$. By then, $S_{h}$ is attacked successfully.

\subsection{Dense Category-wise Attack (DCA)}

It is interesting to investigate our optimization problem Eq.~\ref{categorywiseoptimization} for $p=\infty$.
In this case, our adversarial perturbation generation procedure is based on PGD~\cite{madry2017towards} and is called \emph{Dense Category-wise Attack} (DCA) since it generates dense perturbations compared with SCA. Note that our DCA framework can also be based on other adversarial attacks, e.g., FGSM~\cite{goodfellow2014explaining}. We propose three methods based on DCA, i.e., DCA-G, DCA-L, and DCA-S, depending on the targeted region. They are described in details next. 

\renewcommand{\algorithmicrequire}{\textbf{Input:}}
\renewcommand{\algorithmicensure}{\textbf{Output:}}
\begin{algorithm}[t]
    \setstretch{1.3}
    \caption{Dense Category-wise Attack (DCA)}
    \label{dca}
    \begin{algorithmic}

        \Require
        image $\x$,  $\mathcal{S}$,
         $\mathcal{C}$, $\epsilon$, $max\_iter$, $\mathcal{T}$
        \Ensure
        adversarial example $\x^*$, perturbation $\rb$

        \State{\textbf{Initialize:} $\x_1 \leftarrow \x, p \leftarrow 1$}

        \While{$\mathcal{S} \not\in \varnothing$ and $p \leq max\_iter$}

            \State$G \leftarrow 0, j \leftarrow 1$

            \While{$j \leq k$}

                \If{$S_{\text{j}} \not= \varnothing$}

                    \State $\gb_j\!=\!\frac{\nabla_{\x_p}\sum_{s\in S_{\text{j}}}\text{CE}~(f(\x_p, s),~C_j)}{\|\nabla_{\x_p}\sum_{s\in S_{\text{j}}}\text{CE}~(f(\x_p, s),~C_j)\|_{\infty}}$, \ $G \leftarrow G + \gb_j$

                \EndIf

                \State$j \leftarrow j + 1$

            \EndWhile
            \State $\text{GP} \leftarrow \frac{\epsilon}{max\_iter}\cdot sign(G)$
            \If{DCA-G}
                \State $\x_{p+1} \leftarrow \x_p + \text{GP}$ ~ {\footnotesize/*Refer to Eq.~\ref{eq:global_pert}*/}
             \ElsIf{DCA-L}
                \State $\x_{p+1} \leftarrow \x_p + \text{GP}*\mask^L$ ~ {\footnotesize/*Refer to Eq.~\ref{eq:local_pert}*/}
              \ElsIf{DCA-S}
                \State $\x_{p+1} \leftarrow \x_p + \text{GP}*\mask^S$ ~ {\footnotesize/*Refer to Eq.~\ref{eq:semantic_pert}*/}
              \EndIf
              
            \For{$S_i$ in $\mathcal{S}$}
                \State $S_{i} =$ \text{RemovePixels}($\x_p, \x_{p+1}, S_{i}, \mathcal{T}$)~{\footnotesize/*Refer to Eq.~\ref{eq:update_target}*/}
            \EndFor
            \State$p \leftarrow p + 1$

        \EndWhile

        \State{\textbf{return} $\x^*=\x_p$, $\rb = \x_p - \x$}

    \end{algorithmic}
    \label{algo: algorithm_dca}
\end{algorithm}

\subsubsection{DCA on Global Region (DCA-G)}
Given an input image $\x$ and category-wise target pixel sets $\mathcal{S}$, DCA applies two iterative loops to generate adversarial perturbations.
In each inner loop iteration $j$, 
DCA first computes the total loss of all pixels in target pixel set $S_j$ corresponding to each available category $C_j$ with $\sum_{s\in S_{\text{j}}}\text{CE}~(f(\x, s),~C_j)$,
where CE($\cdot$,$\cdot$) is the traditional cross-entropy loss. Then, it computes local adversarial gradient $\gb_j$ of the loss with respect to the current image $\x$ and normalize it with $L_\infty$ norm:
\begin{equation*}
   \begin{aligned}
       \gb_j =\frac{\nabla_{\x} \sum_{s\in S_{\text{j}}}\text{CE}~(f(\x, s),~C_j)}{\|\nabla_{\x} \sum_{s\in S_{\text{j}}}\text{CE}~(f(\x, s),~C_j)\|_{\infty}}.
\end{aligned}
\label{eq:ad_grad}
\end{equation*}

After that, DCA adds up all $\gb_j$ to generate a total adversarial gradient $G$. In an outer loop iteration $p$, DCA computes the global perturbation (GP) by applying $sign$ operation to the total adversarial gradient $G$ \cite{madry2017towards} as follows,
\begin{equation}
  \begin{aligned}
      \text{GP} = \frac{\epsilon}{max\_iter}\cdot sign(G),
    \end{aligned}
\label{eq:global_pert}
\end{equation}
where $max\_iter$ denotes the maximum number of iterations of the outer loop and the term $\frac{\epsilon}{max\_iter}$ is perturbation size in each iteration \cite{goodfellow2014explaining}. At the end of the outer loop, DCA uses \textit{RemovePixels} of Eq.~\ref{eq:update_target} to remove from $\mathcal{S}$ the target pixels that have already been attacked successfully on the perturbed image.
Since DCA works on the global (whole) region in the original image, we call this method, DCA-G, in short. The pseudo-code of DCA-G is described in Algorithm~\ref{algo: algorithm_dca}.


\subsubsection{DCA on Local Region (DCA-L)} 
As we discussed before, runner-up pixels are usually located near keypoints in the heatmap.  
They are the most important pixels for the object. 
In addition, perturbations in the background from an image may not impact the detection results since they are not related to the objects.  To attack these objects, we only need to attack these significant pixels (i.e., runner-up and keypoints). 

To fulfil the goal, we can use attack mask $\mask^L$ to restrict perturbation around detected objects in order to disallow perturbation in the background. 
Attack mask $\mask^L$ is generated from $\mathcal{S}$ with the following process, called the \textit{GenerateMask} procedure. $\mask^L$ is initialized with a zero matrix of the same size as the input image. We locate each pixel point $s \in S_{i}, \forall i\in[k]$, on the input image and set the same location point of $\mask^L$ to be $1$. After processing all $s$, for each pixel $s=1$ in $\mask^L$, we set all the points in the square box centered at $s$'s with the size of radius (side length) $R^*$ to be $1$ too. The resulting $\mask^L$ is used as the local attack mask. The attack performance depends on the value of $R^*$. We will quantitatively analyze this dependence in the experimental studies reported in Section~\ref{sect::hyperparas}. 

A local perturbation (LP) is obtained by applying $\mask^L$ on the global perturbation (GP):
\begin{equation}
    \begin{aligned}
         \text{LP} = \text{GP} * \mask^L.
    \end{aligned}
\label{eq:local_pert}
\end{equation}
After generating LP, we update $\mathcal{S}$ by removing the points that have been attacked successfully with \textit{RemovePixels} (Eq. ~\ref{eq:update_target}). The pseudo-code of DCA-L is shown in Algorithm~\ref{algo: algorithm_dca}.

\subsubsection{DCA on Semantic Region (DCA-S)} 
\begin{figure}[t]
    \begin{center}
    \includegraphics[width=0.47\textwidth]{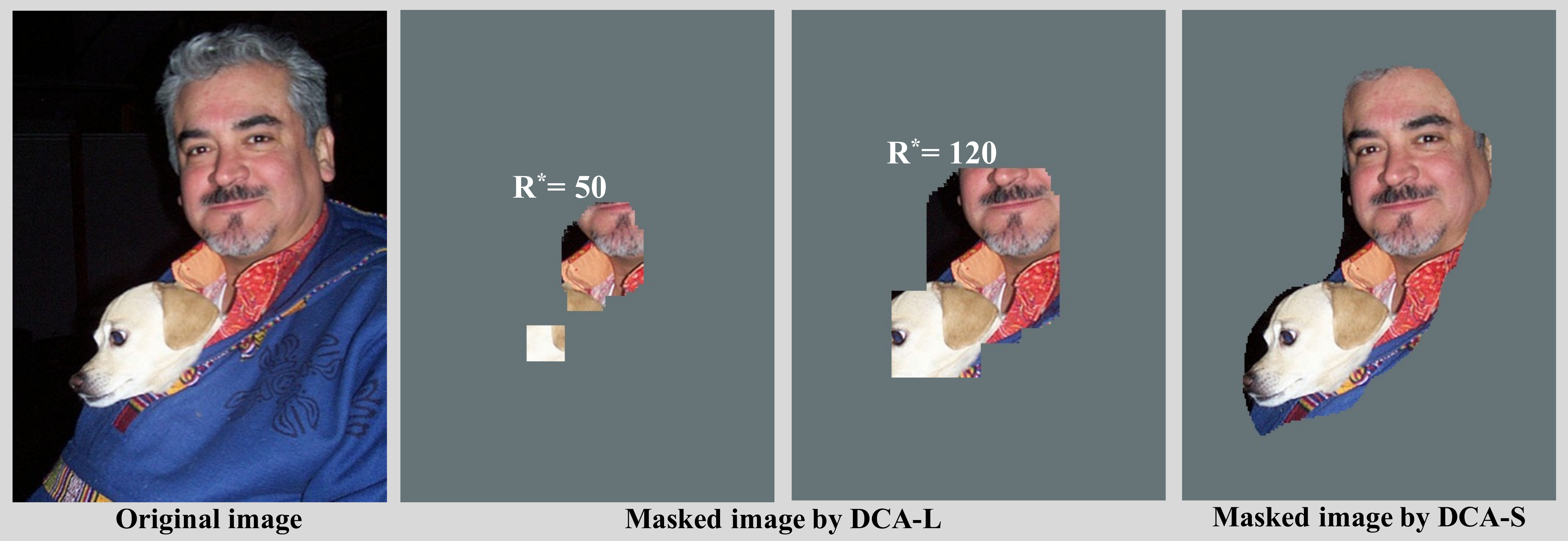}
    \end{center}
    \vspace*{-0.2cm}
    \caption{\small Comparison of masks from DCA-L and DCA-S.
    A proper radius $R^*$ of the DCA-L mask is hard to be determined. A undersized $R^*$ (i.e., $R^*\!\!=\!\!50$) will miss some important regions (e.g., human and dog faces) while an oversized $R^*$ (i.e., $R^*\!\!=\!\!120$) will introduce more useless local perturbations (e.g., perturbations on background), which may lead to worse results. In contrast, the mask generated by DCA-S is more precise. 
    }
    \label{mask_compare}
    \vspace*{-0.7cm}
\end{figure}

Using DCA on the local region may not get a perfect perturbation around objects because run-up pixels may not be all around objects since some of them may not represent semantic information of objects. 
In addition, DCA-L uses a regular square mask around each pixel in $S_, \forall i\in[k]$, which may not precisely select important perturbation signals from the global perturbation, as Fig.~\ref{mask_compare} shows. To address this issue, we propose the DCA-S method that applies DCA to the semantic region of the image.

Since convolutional layers of the CNN-based anchor-free model contain abundant information, especially the spatial information \cite{zhou2016learning, selvaraju2017grad}, which can indicate which regions of an input image have a higher response to the output of the classifier. 
This characteristic demonstrates that the generated perturbation will be more effective if we only attack pixels that are related to the semantic region of objects.
A natural strategy for getting this semantic region is to extract the feature map based on the convolutional layer outputs~\cite{Liu2018pannet,huang2016densenet,Li201fssd}. Shallower layers usually keep much spatial information, leading to a sparse and discontinuous high response region in the feature map. On the other hand, deeper layers contain more global information due to their larger receptive fields, resulting in a large and continuous high response region. 

In prior work \cite{selvaraju2017grad}, the last convolutional layer is used to adopt the key region. Discarding spatial information from shallower layers makes this method hard to accurately target locations of small objects.
Inspired by~\cite{huang2016densenet,Liu2018pannet,Li201fssd}, we propose a multi-layer semantic information region selection (MSIRE) method, to be described in detail next, to extract and integrate semantic information from both deep and shallow layers. 
By integrating the gradient information from these layers, we can construct a semantic mask that contains the most informative semantic region of objects, which is then combined with the global perturbation (GP) to generate the final perturbation.

Now we describe the MSIRE method.  
Let $L=\{l_{i}\}_{i=1}^{n}$ be the set of layers that we consider to extract key region, 
where  $l_{i}$ is the $i$-th layer containing feature map activations $h_i$. 
In our experiments, $n$ is set to 4.
For layer $l_i$, we calculate and sum the gradient of the score for category $C_j$ of all pixels in target pixel set $S_j$ with respect to its feature map activation $h_i$, and sum them up across all categories, 
\begin{equation*}
   \begin{aligned}
        G_{i}=\sum_{j\in [k]}\frac{\partial \sum_{s\in S_{\text{j}}}f_j(\x, s)}{\partial h_i}
   \end{aligned}
\end{equation*}
Then normalize gradient $G_i$ to [0,1] and upsample to the size of image $\x$:
\begin{equation*}
   \begin{aligned}
        \hat{G}_{i}=\Phi\Big(\frac{G_i-\min(G_i)}{\max(G_i)-\min(G_i)}\Big),⁡
   \end{aligned}
\end{equation*}
where $\Phi(\cdot)$ denotes the operation of upsampling and $\min(G_i)$ and $\max(G_i)$ are the minimal and maximal value of elements in $G_i$, respectively. The mask corresponding to layer $l_i$ can be obtained using a threshold $\mathcal{T}_{s}$:
\begin{equation*}
   \begin{aligned}
        mask_i=\delta(\hat{G}_{i}>\mathcal{T}_{s}),⁡
   \end{aligned}
\end{equation*}
where $\delta(\cdot)$ is an impulse response that turns a pixel greater than $\mathcal{T}_{s}$ to 1, otherwise to 0. Combining $mask_i, \forall i\in[n]$ as
\begin{equation*}
   \begin{aligned}
        \mask^S=\prod_{i=1}^n mask_i,⁡
   \end{aligned}
\end{equation*}
where $\mask^S$~denotes the final semantic region attack mask.

After obtaining $\mask^S$, we combine it with the global perturbation (GP) to get a semantic region guided perturbation (SP) as follows,
\begin{equation}
    \begin{aligned}
         \text{SP} = \text{GP} * \mask^S.
    \end{aligned}
\label{eq:semantic_pert}
\end{equation}
Like its siblings, after generating the perturbation, DCA-S updates $\mathcal{S}$ with \textit{RemovePixels} (Eq.~\ref{eq:update_target}). The pseudo-code of DCA-S is shown in Algorithm~\ref{algo: algorithm_dca}.

\section{Experiments} \label{sec:experiments}
In this section, we will evaluate the performance of the proposed adversarial attack (\textit{i.e.}, SCA, DCA-G, DCA-L, and DCA-S) for both object detection and human pose estimation based on anchor-free detector CenterNet~\cite{zhou2019centernet}.

\subsection{Experimental Settings}

\subsubsection{Detectors and Datasets}
For \underline{object detection}, we evaluate our proposed attack on two public datasets: PascalVOC \cite{everingham2015pascal} and MS-COCO \cite{lin2014microsoftcoco}. 
We use the two pre-trained detectors (CenterNet with two different backbones: ResNet18 (R18) \cite{he2016resnet} and DLA34 \cite{yu2018dla}) from \cite{zhou2019centernet} in our experiments. These two detectors are pre-trained on the training set of PascalVOC (including PascalVOC 2007 and PascalVOC 2012) and MS-COCO 2017, respectively. 
Our adversarial examples are generated on the test set of PascalVOC 2007, which consists of 4,592 images and 20 categories, and the validation set of MS-COCO 2017, which contains 5,000 images and 80 object categories.
For \underline{human pose estimation}, we use the pre-trained detector (CenterNet with DLA34 \cite{yu2018dla} backbone) from \cite{zhou2019centernet} that is trained on the training set of MS-COCO Keypoints 2017. Our adversarial examples are generated on its validation set, which contains 5,000 images and 17 categories of keypoints for humans. 


\subsubsection{Evaluation Metrics}
We evaluate the performance of both white-box and black-box attacks with the following metrics. 
\begin{itemize}
    \item The attack performance is evaluated by computing the decreased percentage of mean average precision (mAP), referred to as the \textbf{mAP Score Degradation Ratio (ASR)}, which is defined as, 
    \begin{equation*}
       \begin{aligned}
           \text{ASR}=1-\frac{\text{mAP}_{attack}}{\text{mAP}_{clean}},
       \end{aligned}
       \label{eqasr}
    \end{equation*}
    where mAP$_{attack}$ denotes the mAP of the targeted object detector on adversarial examples, and mAP$_{clean}$ denotes the mAP on clean samples. Higher ASR means better white-box attack performance.

    \item We also use the $L_0$ and $L_2$ norms of perturbation $\rb$, $P_{L_0}$ and $P_{L_2}$,     respectively, to quantify the \textbf{perceptibility} of the adversarial perturbation. $P_{L_0} = \|\rb\|_0$ quantifies the proportion of perturbed pixels. A lower value means that fewer image pixels are perturbed. For $P_{L_2} = \|\rb\|_2$, a greater value usually signifies a more perceptible perturbation to humans. 
    
    \item For black-box attacks, transferability of adversarial examples generated with other detectors and tested with the target detector is used to measure the attack performance. More specifically, the attack performance of block-box attacks is measured by the ratio, referred to as the \textbf{Attack Transfer Ratio (ATR)}, of the ASR on the target model, ASR$_{target}$, to the ASR on the generating model, ASR$_{origin}$: 
    \begin{equation*}
       \begin{aligned}
            \text{ATR}=\frac{\text{ASR}_{target}}{\text{ASR}_{origin}},
       \end{aligned}
       \label{eqatr}
    \end{equation*}
    Higher ATR means better transferability.
    
    In our experiments, for \underline{object detection}, black-box adversarial examples are generated on CenterNet with one backbone (R18 or DLA34) and tested on CenterNet with a different backbone (DLA34, R18, and ResNet101 (R101)). We also test these adversarial examples on other detectors, including anchor-free (CornerNet~\cite{law2018cornernet} with backbone Hourglass~\cite{newell2016stacked}) and anchor-based detectors (Faster-RCNN~\cite{ren2015fasterrcnn} and SSD~\cite{liu2016ssd}). For \underline{human pose estimation}, all adversarial examples are generated on CenterNet with backbone DLA34 and tested on CenterNet with backbone Hourglass. 
    
    Additionally, we follow \cite{dziugaite2016study} to simulate a real-world attack transferring scenario, wherein generated adversarial examples are saved in the JPEG format and then reloaded to attack the target model. 
    In real-world applications, images are usually saved in a compression format. JPEG is a most commonly used lossy compression standard for images. The transferability test in this way requires adversarial examples to be robust to the JPEG compression like in most real-world applications.    

\end{itemize}

\subsubsection{Comparison Methods and Implementation Details} 

For \underline{object detection}, since there is no existing attack dedicated to anchor-free detectors, we use an existing state-of-the-art attack designed for attacking anchor-based detectors, i.e., DAG~\cite{xie2017dag} with VGG16~\cite{simonyan2014very} backbone for attacking Faster-RCNN (FR) as the attack to compare with. 
For \underline{human pose estimation}, we compare with existing methods proposed in~\cite{jain2019robustness} for attacking human pose estimation systems, called FHPE and PHPE, which are based on FGSM and PGD, respectively. 

Our methods are implemented with Python 3.6 and Pytorch 1.1.0. All the experiments are conducted with an Intel Core i9-7960 and an Nvidia GeForce GTX-1070Ti GPU. We set \textit{max\_iter\_outer}, \textit{max\_iter\_inner}, and \textit{max\_iter} to 50, 20, and 30, respectively. 
The default values of $R^*$, $\mathcal{T}_s$, $\mathcal{T}$,  and $\epsilon$ are 60, 0.5, 0.1, and 5\% of the maximum value of the pixels in an image, respectively. 
All input images are resized to 512$\times$512.



\subsection{Experimental Results on Object Detection}

\subsubsection{White-Box Attack Performance}

\begin{table}[t]
\centering
\label{overall_white}
\resizebox{1.0\columnwidth}{!}{
    \begin{tabular}{c|c|c||c|c|c|c|c|c}
    \hline
                Data    & Method& Network         & mAP$_{clean}$       & mAP$_{attack}$     & ASR  & $P_{L_2}$ ($\times 10^{-3}$)  & $P_{L_0}$    & Time (s)             \\ \hline \hline
    \multirow{10}{*}{\rotatebox[origin=c]{90}{PascalVOC}}&DAG & FR & 0.70 & 0.050 & 0.92 & 3.20 & 0.990 & 9.8  \\ \cline{2-9}
    & SCA & R18 & 0.71 & 0.060 & 0.91 & \textbf{0.41} & \textbf{0.002} & 32.5  \\ 
    &SCA & DLA34 & 0.79 & 0.110 & 0.86 & 0.44 & 0.003 & 117.3  \\ 
    &DCA-G & R18 & 0.71 & 0.070 & 0.90 & 5.20 & 0.990 & \textbf{0.7} \\
    &DCA-G & DLA34 & 0.79 & 0.050 & \textbf{0.94} & 5.10 & 0.990 & 1.2 \\ 
    &DCA-L & R18 & 0.71 & 0.080 & 0.89 & 2.70 & 0.320 & 1.6 \\
    &DCA-L & DLA34 & 0.79 & 0.080 & 0.90 & 2.80 & 0.320 & 3.1 \\
    &DCA-S & R18 & 0.71 & 0.070 & 0.90 & 2.40 & 0.260 & 2.2 \\
    &DCA-S & DLA34 & 0.79 & 0.060 & 0.92 & 2.20 & 0.280 & 4.0 \\\hline\hline
 
    \multirow{8}{*}{\rotatebox[origin=c]{90}{MS-COCO}}
    & DAG & FR   & 0.35 & 0.040 & 0.89 & 5.00 & 0.990 & 20.4\\ \cline{2-9}
    & SCA & R18   & 0.28 & 0.027 & 0.91 & \textbf{0.48} & \textbf{0.004} & 50.4\\ 
    & SCA & DLA34 & 0.37 & 0.030 & 0.92 & 0.49 & 0.007 & 216.0\\ 
    & DCA-G & R18 & 0.28 & 0.002 & \textbf{0.99} & 5.80 & 0.990 & \textbf{2.4}\\ 
    & DCA-G & DLA34 & 0.37 & 0.002 & \textbf{0.99} & 5.90 & 0.990 & 4.9  \\ 
    & DCA-L & R18 & 0.28 & 0.006 & 0.98 & 3.20 & 0.380 & 3.4 \\
    & DCA-L & DLA34 & 0.37 & 0.008 & 0.98 & 3.30 & 0.390 & 6.3 \\
    & DCA-S & R18 & 0.28& 0.003 & \textbf{0.99} & 2.80 & 0.300 & 4.7 \\
    & DCA-S & DLA34 & 0.37 & 0.004 & \textbf{0.99} & 3.00 & 0.310 & 8.5 \\\hline
    \end{tabular}
}
\vspace{0.2em}
\caption{\small White-box performance comparison. ``Time'' is the average time to generate an adversarial example.  The best results
are shown in bold.
}
\label{table_overalwhite_icme}
\vspace*{-0.5cm}
\end{table}

\begin{figure*}[t]
    \centering
 \includegraphics[width=1.0\linewidth]{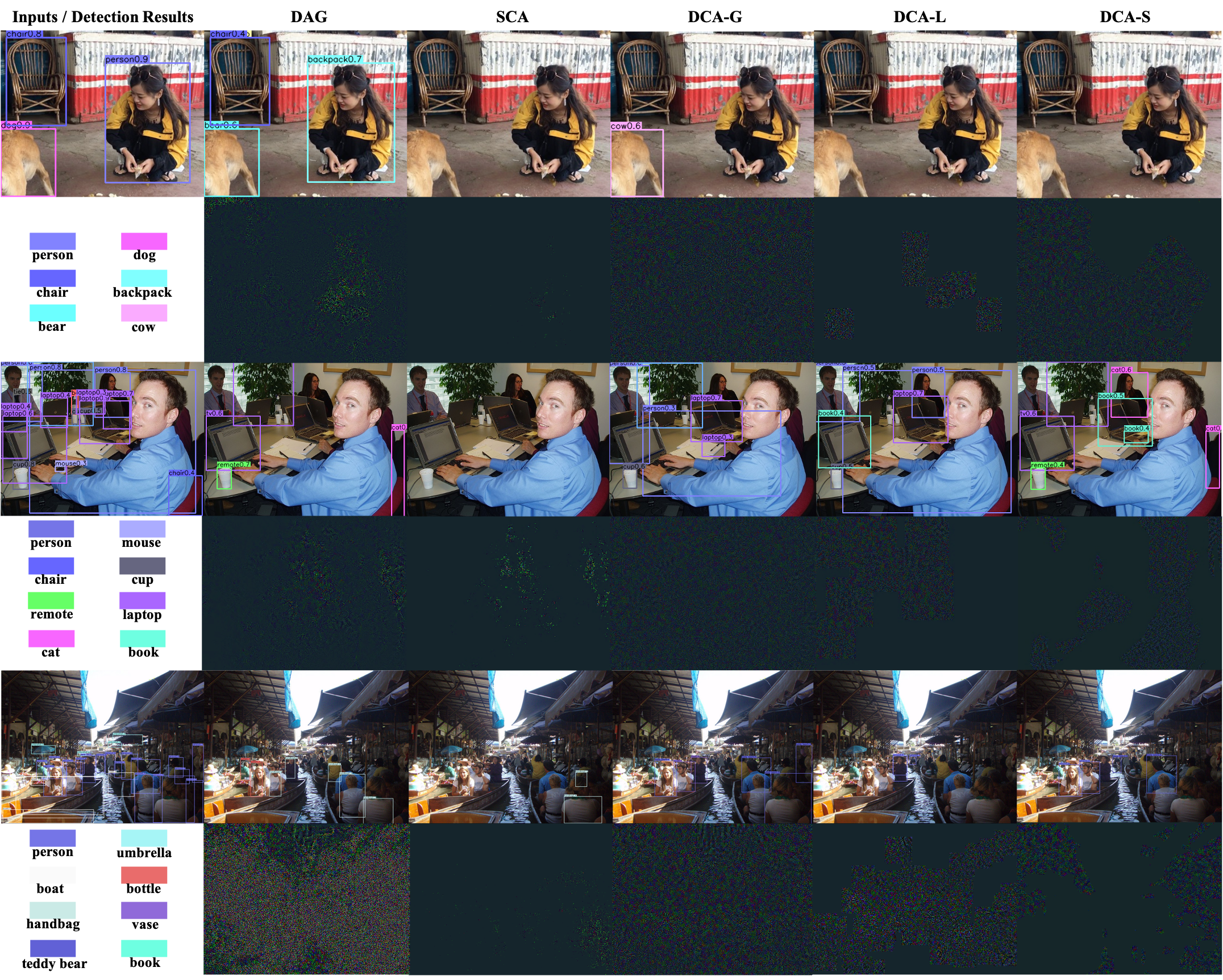}
    \vspace{-0.4cm}
    \caption{\small
    Qualitative comparison between DAG and our methods on the object detection task. Three examples are presented. \textbf{Column 1:} Detection results of clean inputs. \textbf{Column 2:} DAG's attack results and perturbations. \textbf{Column 3:} SCA's attack results and perturbations.
    \textbf{Columns 4 - 6:} the attack results and perturbations of DCA-G, DCA-L, and DCA-S, respectively.
    The perturbations are magnified by a factor of 30 for better visibility.}
\label{fig: qualitative_obj}
\vspace{-0.5cm}
\end{figure*}

\begin{figure}[t]
 \begin{center}
    \includegraphics[width=0.47\textwidth]{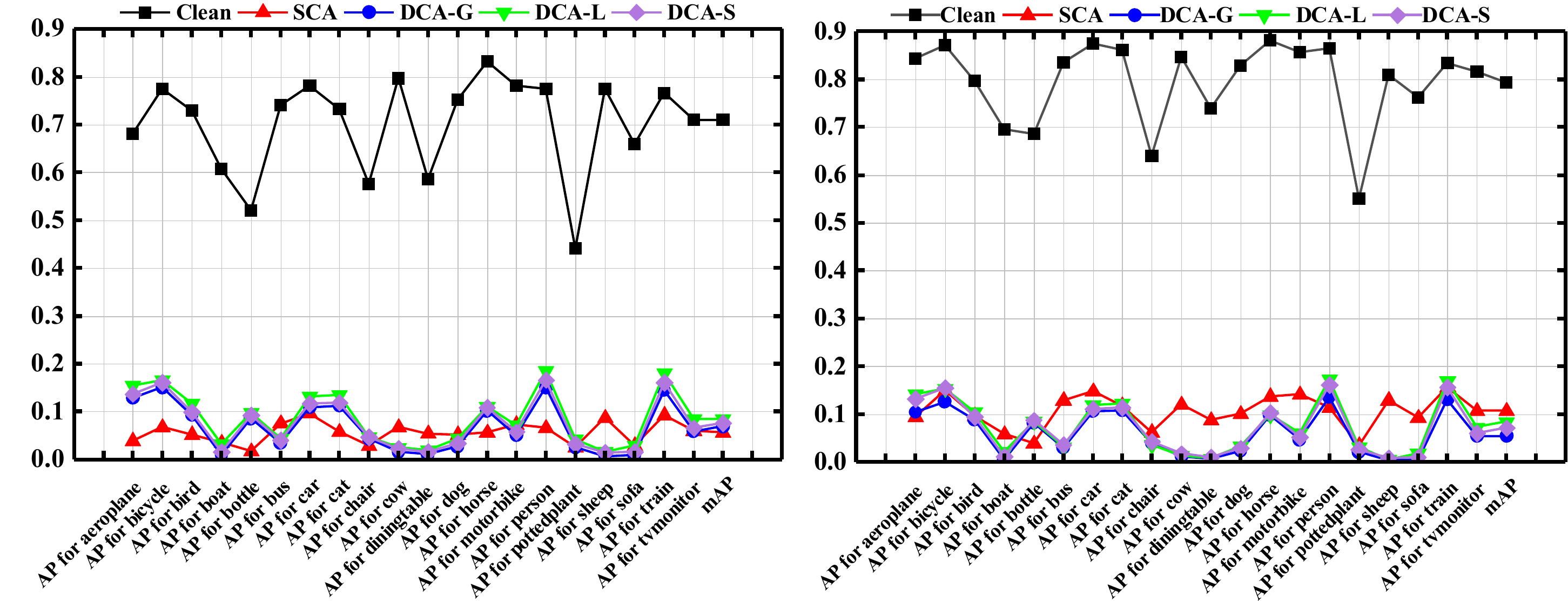}
    \end{center}
    \vspace{-0.2cm}
    \caption{\small The AP of each object category on clean inputs and adversarial examples generated by SCA and DCA-methods on PascalVOC with CenterNet using R18 (\textbf{left}) and DLA34 (\textbf{right}) backbones on PascalVOC.}
   \label{pascal_detail}
   \vspace{-0.5cm}
\end{figure}


\begin{figure}[t]
    \begin{center}
    \includegraphics[width=0.49\textwidth]{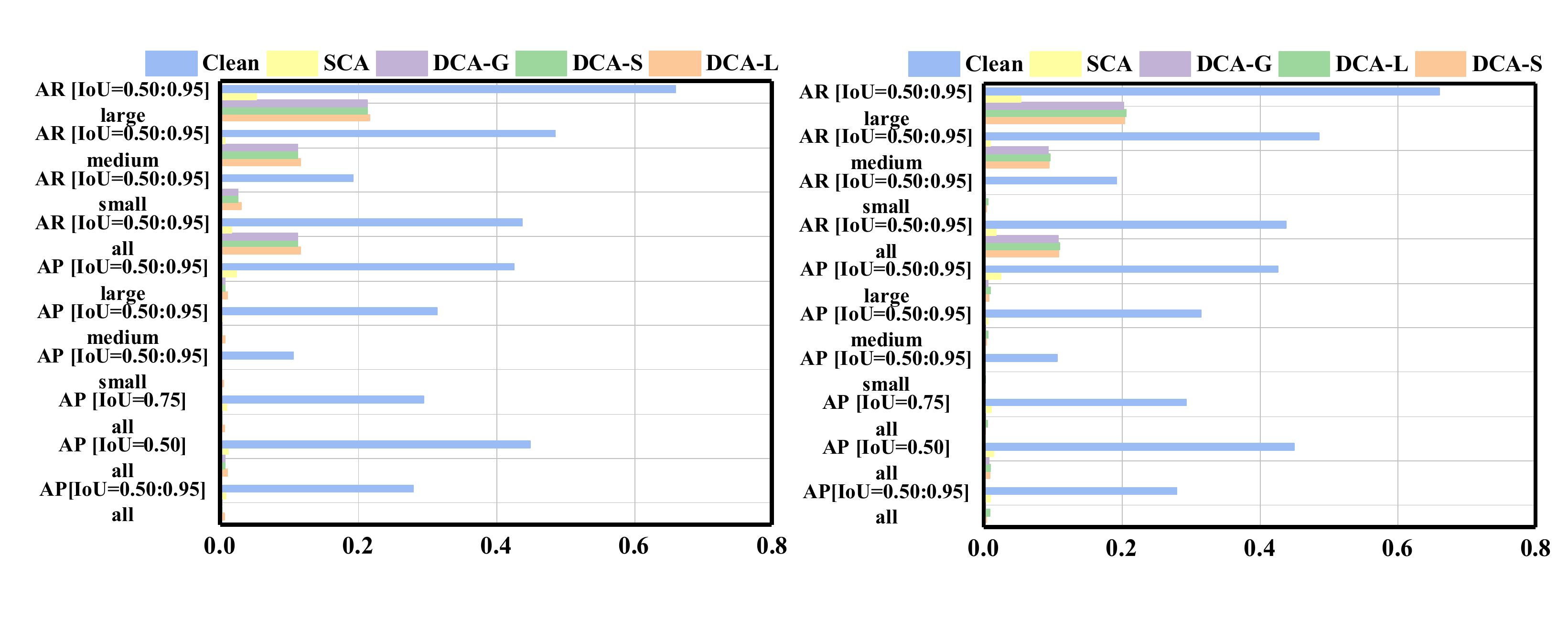}
    \end{center}
    \vspace{-0.4cm}
    \caption{\small The AR (\textbf{top four}) and AP (\textbf{bottom six}) performance of different sizes of objects on clean inputs and adversarial examples generated SCA and DCA-based methods on MS-COCO with CenterNet using R18 (\textbf{left}) and DLA34 (\textbf{right}) backbones.}
   \label{coco_detail}
   \vspace{-0.5cm}
\end{figure}


Table~\ref{table_overalwhite_icme} shows the white-box attack results on both PascalVOC and MS-COCO. 
\begin{itemize}
    \item \textit{For PascalVOC}, we can see that 
    DCA-G with the DLA34 backbone achieves the best ASR. SCA with both backbones produces much smaller perturbations than other methods and DCA-G with the R18 is 14 times faster than DAG. Furthermore, while the ASR performance of both DCA-L and DCA-S are very close to that of  DAG, DCA-L and DCA-S produce smaller perturbations and have lower time complexity than DAG in generating adversarial examples.
    \item \textit{For MS-COCO}, 
    both DCA-G and DCA-S achieve the highest ASR, 99.0\%, which is significantly higher than that of DAG. Like on PascalVOC, the ASR performance of SCA with both R18 and DLA34 is in the same ballpark as that of DAG,  
    SCA produces much smaller perturbations than other methods in terms of both $P_{L_2}$ and $P_{L_0}$, 
    and DCA-G is the fastest in generating an adversarial example.
\end{itemize}
In general, both SCA and DCA achieve state-of-the-art attack performance. Specifically, DCA can achieve high ASR and low time complexity, while SCA can produce much smaller perturbations without degrading ASR much.  We can see that $\mathop{P_{L_0}}$ of SCA is lower than 1\%, implying that SCA can successfully attack detectors by perturbing only a small percentage of pixels of the original image.
Comparing DCA-L and DCA-G, while it may slightly decrease the performance of ASR, DCA-L using a local region significantly decreases the size of perturbation. Comparing DCA-S and DCA-L, DCA-S not only improves the performance of ASR but also reduces the perturbation size in terms of both $P_{L_2}$ and $P_{L_0}$. This implies that the DCA on the semantic region is better than it on the local region. A qualitative comparison between DAG and our methods is shown in Fig.~\ref{fig: qualitative_obj}. It is clear that perturbations generated by SCA and DCA-based methods are hard to be perceived by humans. 


We also show in Fig. \ref{pascal_detail} the Average Precision (AP) of each object category on clean inputs and adversarial examples generated by our methods on PascalVOC with Centernet using both R18 and DLA34 backbones. The AP drops by a roughly similar percentage for all the object categories. Fig.~\ref{coco_detail} shows the AP and Average Recall (AR) of SCA and DCA-based methods on MS-COCO with CenterNet. We can notice that small objects are more vulnerable to adversarial examples than bigger ones. One possible explanation is that bigger objects usually have more keypoints than smaller objects on the heatmap and our algorithms need to attack all of them.

\subsubsection{Black-Box Attack Performance}


\begin{table}[t]
    \begin{center}
    \label{tab_black_pascal}
    \resizebox{1.0\columnwidth}{!}{
        \begin{tabular}{c||c|c||c|c||c|c||c|c||c|c}
            \hline
            \multirow{2}{*}{\diagbox{\textbf{From}}{\textbf{To}}} & \multicolumn{2}{c||}{R18} & \multicolumn{2}{c||}{DLA34} & \multicolumn{2}{c||}{R101} & \multicolumn{2}{c||}{Faster-RCNN} & \multicolumn{2}{c}{SSD} \\ \cline{2-11} 
                                   & mAP           & ATR       & mAP         & ATR      & mAP            & ATR           & mAP            & ATR           & mAP           & ATR           \\ \hline \hline
            Clean & 0.71 & -- & 0.77 & -- & 0.79 & -- & 0.71 & -- & 0.74 & --\\ \hline
            DAG & 0.65 & 0.09 & 0.75 & 0.03 & 0.72 & 0.10 & 0.05 & -- & 0.72 & 0.03\\ \hline
            R18-SCA & 0.06 & -- & 0.62 & 0.21 & 0.61 & 0.25 & 0.55 & 0.25 & 0.70 & 0.10\\ \hline
            DLA34-SCA & 0.42 & \textbf{0.47} & 0.11 & -- & 0.53 & \textbf{0.38} & 0.44 & \textbf{0.44} & 0.62 & \textbf{0.19} \\ \hline
            R18-DCA-G & 0.07 & -- & 0.62 & 0.22 & 0.65 & 0.20 & 0.61 & 0.16 & 0.72 & 0.03\\ \hline
            DLA34-DCA-G & 0.50 & 0.31 & 0.05 & -- & 0.62 & 0.23 & 0.53 & 0.27 & 0.67 & 0.10\\ \hline
            R18-DCA-L & 0.08 & -- & 0.61 & \textbf{0.23} & 0.63 & 0.23 & 0.55 & 0.25 & 0.68 & 0.09\\ \hline
            DLA34-DCA-L & 0.48 & 0.35 & 0.06 & -- & 0.60 & 0.24 & 0.51 & 0.37 & 0.66 & 0.12\\ \hline
            R18-DCA-S & 0.07 & -- & 0.65 & 0.17 & 0.65 & 0.17 & 0.61 & 0.16 & 0.73 & 0.01\\ \hline
            DLA34-DCA-S & 0.52 & 0.29 & 0.06 & -- & 0.62 & 0.22 & 0.57 & 0.21 & 0.70 & 0.06\\ \hline
            \end{tabular}}
    \end{center}
\vspace{-0.2cm}
\caption{\small Black-box attack results on the PascalVOC dataset. \textbf{From:} the leftmost column denotes the models which adversarial examples are generated from.  \textbf{To:} the top row denotes the target models that adversarial examples transfer to (i.e., are tested on). ``--'' means the value is unavailable (not defined).}
\label{table_blackpascal_icme}
\vspace{-0.1cm}
\end{table}

\begin{table}[t]
    \begin{center}
    \label{tab_black_coco}
    \vspace{-0.25cm}
    \resizebox{1.0\columnwidth}{!}{
        \begin{tabular}{c||c|c||c|c||c|c||c|c}
            \hline
            \multirow{2}{*}{\diagbox{\textbf{From}}{\textbf{To}}} & \multicolumn{2}{c||}{R18} & \multicolumn{2}{c||}{DLA34} & \multicolumn{2}{c||}{R101} & \multicolumn{2}{c}{CornerNet} \\ \cline{2-9} 
                                  & mAP           & ATR        & mAP       
                                  & ATR      & mAP            & ATR           & mAP            & ATR            \\ \hline \hline
            Clean & 0.29 & -- & 0.37 & -- & 0.37 & -- & 0.43 & --\\ \hline
            DAG & 0.24 & 0.19 & 0.33 & 0.12 & 0.31 & 0.18 & 0.40 & 0.08 \\ \hline
            R18-SCA & 0.02 & -- & 0.27 & \textbf{0.30} & 0.24 & 0.39 & 0.35 & 0.20\\ \hline
            DLA34-SCA & 0.07 & \textbf{0.82} & 0.03 & -- & 0.09 & \textbf{0.82} & 0.12 & \textbf{0.78}  \\ \hline
            R18-DCA-G & 0.00 & -- & 0.29 & 0.21 & 0.28 & 0.25 & 0.38 & 0.12\\ \hline
            DLA34-DCA-G & 0.10 & 0.67 & 0.00 & -- & 0.12 & 0.69 & 0.13 & 0.72\\ \hline
            R18-DCA-L & 0.01 & -- & 0.27 & 0.28 & 0.27 & 0.28 & 0.36 & 0.17\\ \hline
            DLA34-DCA-L &  0.10 & 0.67 & 0.01 & -- & 0.12 & 0.69 & 0.13 & 0.71\\ \hline
            R18-DCA-S & 0.00 & -- & 0.31 & 0.16 & 0.29 & 0.22 & 0.39 & 0.09\\ \hline
            DLA34-DCA-S & 0.12 & 0.59 & 0.00 & -- & 0.13 & 0.64 & 0.13 & 0.72\\ \hline
        \end{tabular}}
    \end{center}
\vspace{-0.2cm}
\caption{\small Black-box attack results on the MS-COCO dataset. \textbf{From:} the leftmost column denotes the models which adversarial examples are generated from.  \textbf{To:} the top row denotes the target models that adversarial examples transfer to (i.e., are tested on). ``--'' means the value is unavailable (not defined).}
\label{table_blackcoco_icme}
\vspace{-0.5cm}
\end{table}

\begin{figure*}[t]
    \centering
    \includegraphics[width=1\linewidth]{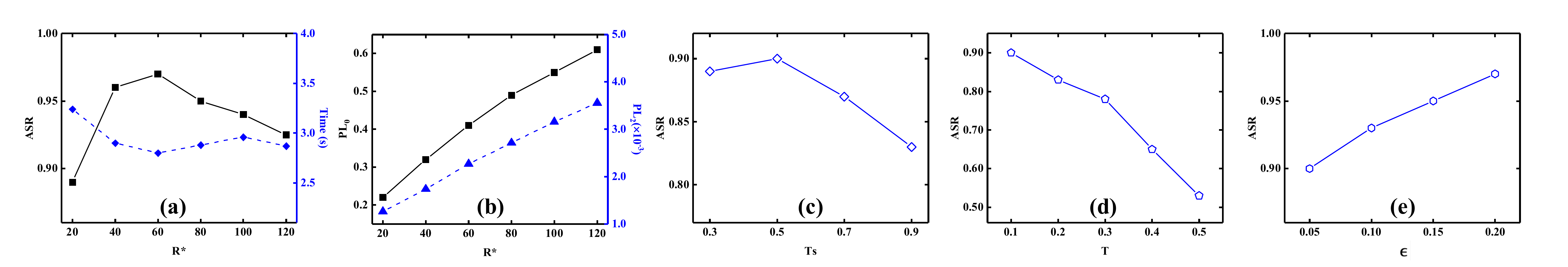}
    \vspace{-0.5cm}
    \caption{\small Sensitivity analysis of hyperparameters. }
    \label{ar}
    \vspace{-0.5cm}
\end{figure*}

In evaluating black-box attack performance, adversarial examples are generated with Centernet using the R18 or DLA34 backbone for our proposed methods or with Faster-RCNN for DAG and tested with other object detection models. 
Table~\ref{table_blackpascal_icme} and Table~\ref{table_blackcoco_icme} show the black-box attack performance on PascalVOC and MS-COCO, respectively.

\begin{itemize}
    \item \textit{Attack transferability on PascalVOC}. 
    From Table~\ref{table_blackpascal_icme}, we can see that adversarial examples generated by our methods can successfully transfer to not only CenterNet with different backbones but also completely different types of object detectors such as Faster-RCNN and SSD. 
    Specifically, we can see that SCA with the DLA34 backbone achieves the best ATR and SCA with the R18 backbone achieves similar ATR to the DCA-based methods with both the R18 and DLA34 backbones.
    on the other hand, adversarial examples generated by DAG with Faster-RCNN have 
    a much poorer transferability in attacking CenterNet and SSD than our proposed methods, esp. SCA with the DLA34 backbone. 
    In addition, we can also see that our proposed methods have better transferability in attacking CenterNet and Faster-RCNN than in attacking SSD, which implies that SSD is less reliable on highly informational points in the CenterNet-generated heatmap than Faster-RCNN and CenterNet with a different backbone. 
    
    \item \textit{Attack Transferability on MS-COCO.} From Table~\ref{table_blackcoco_icme}, we can see that adversarial examples generated with our proposed methods with the DLA34 backbone have significantly higher ATR than other methods including our proposed methods with the R18 backbone. Our proposed methods with the DLA34 backbone can attack not only  CenterNet with different backbones but also CornerNet. Like on PascalVOC, DAG has a poor transferability in attacking CenterNet and CornerNet on MS-COCO too. 
\end{itemize}

\subsection{Sensitivity Analysis of Hyperparameters}
\label{sect::hyperparas}

In our algorithms, we have four hyperparameters that need to be tuned. To study their sensitivity, we generate adversarial examples by attacking CenterNet with the R18 backbone on MS-COCO for evaluating $R^*$ and on PascalVOC for evaluating $\mathcal{T}_s$, $\mathcal{T}$, and $\epsilon$. Then we report the performance with respect to each hyperparameter. 
More details are provided as follows.

\subsubsection{Sensitivity Analysis of $R^*$} In DCA-L, we need to use $R^*$ to control the size of extracted local regions. It is clear that $R^*$ correlates with the attack performance ASR, $P_{L_0}$, $P_{L_2}$, and time consumption. We show these relationships in Fig.~\ref{ar} (a) and (b). From these figures, we can draw three conclusions. First, the ASR of DCA-L correlates positively with $R^*$ when $R^*$ is less than 60. However, ASR is stable or slightly decreased afterward. The reason is that the oversized mask may introduce more useless perturbation, resulting in a worse effect. Second, $P_{L_0}$ and $P_{L_2}$ of the perturbation correlates positively with $R^*$. This is because higher $R^*$ means bigger attack masks. Finally, the average attack time of DCA-L correlates negatively with $R^*$ when $R^*$ is lower than 48 and becomes stable afterward.

\subsubsection{Sensitivity Analysis of $\mathcal{T}_s$} In DCA-S, hyperparameter $\mathcal{T}_s$ is used to select semantic regions or pixels that contain more informative gradients. Fig.~\ref{ar} (c) shows attack performance ASR with different $\mathcal{T}_s$. We can see that the best ASR occurs when $\mathcal{T}_s$ is 0.5. ASR decreases when $\mathcal{T}_s$ increases from 0.5, which can be explained by the fact that a too large value of $\mathcal{T}_s$ makes the mask too small, resulting in many useful pixels related to objects excluded from the mask and thus degraded attack performance. 

\subsubsection{Sensitivity Analysis of $\mathcal{T}$} The default value of the visual threshold is 0.3 in the Centernet. We use DCA-G to evaluate the sensitivity of $\mathcal{T}$. The results are reported in Fig.~\ref{ar} (d). 
We can see that ASR shows an obvious decreasing trend when $\mathcal{T}$ increases. In particular, the decreasing trend becomes sharper when $\mathcal{T}$ is greater than 0.3, which can be explained that more runner-up points (pixels) are included in the target pixel sets if $\mathcal{T}$ is below 0.3, resulting in improved attacking performance. We can also see that the performance gap is more than 0.1 when $\mathcal{T}$ changes from $0.3$ to $0.1$. This implies that runner-up points have a significant impact on the attack performance.  On the other hand, if $\mathcal{T}$ is higher than 0.3, some keypoints are excluded from the target pixel sets and thus unattacked, leading to faster degraded ASR. 

\subsubsection{Sensitivity Analysis of $\epsilon$} We analyze the sensitivity of the amplitude of the perturbation $\epsilon$ with DCA-G. The results are reported in Fig.~\ref{ar} (e). A large $\epsilon$ means a larger perturbation size. We expect ASR increases with increasing $\epsilon$. This is confirmed by the experimental results shown in the figure.


\subsection{Experimental Results on Human Pose Estimation}

In the MS-COCO Keypoints dataset, there are 17 categories of keypoints for humans. We attack target pixels from each category and report the attacking performance next.

\begin{table}[t]
\centering
\resizebox{1.0\columnwidth}{!}{
    \begin{tabular}{c|c|c||c|c|c|c|c|c}
    \hline
                Data    & Method& Network         & mAP$_{clean}$       & mAP$_{attack}$     & ASR  & $P_{L_2}$ ($\times 10^{-3}$)  & $P_{L_0}$    & Time (s)             \\ \hline \hline
    \multirow{5}{*}{\rotatebox[origin=c]{90}{\makecell{MS-COCO \\ Keypoints}}}&FHPE & DLA34 & 0.53 & 0.31 & 0.42 & 0.33 & 0.990 & \textbf{0.6}  \\ 
    & PHPE & DLA34 & 0.53 & 0.11 & 0.79 & 0.20 & 0.990 & 4.2     \\ \cline{2-9}
    & SCA & DLA34 & 0.53 & 0.03 & 0.93 & \textbf{0.17} & \textbf{0.002} & 130.1  \\ 

    &DCA-G & DLA34 & 0.53 & 0.00 & \textbf{1.00} & 0.62 & 0.990 & 11.2 \\

    &DCA-L & DLA34 & 0.53 & 0.04 & 0.92 & 0.45 & 0.250 & 12.7 \\

    &DCA-S & DLA34 & 0.53 & 0.03 & 0.94 & 0.36 & 0.180 & 15.2\\
\hline
    \end{tabular}
}
\vspace{0.5em}
\caption{\small White-box attack results on the MS-COCO Keypoints dataset. ``Time'' is the average time to generate an adversarial example.}
\label{table_overalwhite_hp}
\end{table}

\subsubsection{White-Box Attack}
The white-box attack results on the MS-COCO Keypoints dataset are summarized in Table~\ref{table_overalwhite_hp}. We can see that our DCA-G has the highest ASR, which is 1. This means DCA-G has attacked all the testing images successfully. Since FHPE and PHPE are based on the conventional FGSM and PGD methods and thus very simple in terms of complexity, we expect them to have less attack time, which is confirmed by the results shown in Table~\ref{table_overalwhite_hp}. However, these two comparison methods have much lower ASR than our proposed methods. All of our proposed methods outperform these two baselines on the ASR metric.

\begin{figure*}[t]
    \centering
  \includegraphics[width=1.0\linewidth]{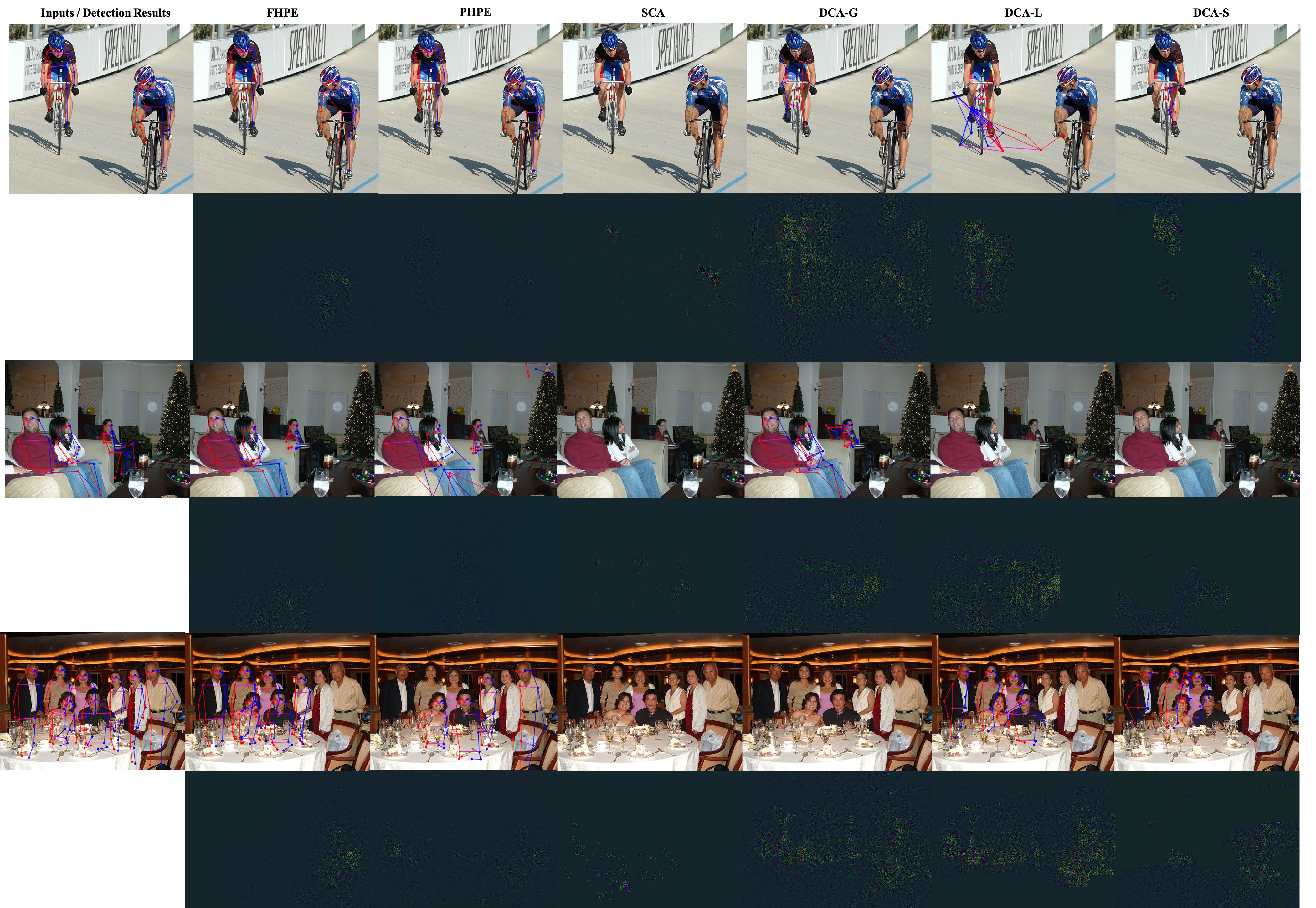}
    \vspace{-0.5cm}
    \caption{\small
    Qualitative comparison between the comparison methods and our proposed methods on the human pose estimation task. Three examples are presented. \textbf{Column 1:} Detection results of clean inputs. \textbf{Column 2:} FHPE attacked results and perturbations. \textbf{Column 3:} PHPE attacked results and perturbations. 
    \textbf{Column 4:} SCA attacked results and  perturbations.
    \textbf{Column 5 - 7:} DCA-G, DCA-L, and DCA-S attacked results and their perturbations, respectively.
    The perturbations are magnified by a factor of 30 for better visibility. }
\label{fig: qualitative_pos}
\end{figure*}

On the other hand,  it is clear that our SCA has the lowest $P_{L0}$ and $P_{L2}$ values, which is consistent with the observation on the object detection task. In addition, our DCA-L and DCA-S also have lower $P_{L0}$ and $P_{L2}$ values than DCA-G, which is due to the role of masks used in both DCA-L and DCA-S. 
A qualitative comparison between the comparison methods and our proposed methods is shown in Fig.~\ref{fig: qualitative_pos}. 

\subsubsection{Black-Box Attack and Transferability}
The black-box attack results are reported in Table~\ref{table_blackcoco}. We can see that our SCA achieves the best transferability, which is the same as on the object detection task. Furthermore, all of our proposed methods outperform the baseline methods, which implies that our category-wise attack can improve the transferability of adversarial examples.

\begin{table}[t]
    \begin{center}
    \label{tab_black_coco}
    \resizebox{1\columnwidth}{!}{
    \begin{tabular}{|cc|ccccccc|}
    \hline
    \multicolumn{2}{|c|}{\multirow{2}{*}{\diagbox{\textbf{To}}{\textbf{From}}}}    & \multicolumn{7}{c|}{DLA34}                                                                                                                            \\ \cline{3-9} 
    \multicolumn{2}{|c|}{}                     & \multicolumn{1}{c|}{Clean} & \multicolumn{1}{c|}{FHPE} & \multicolumn{1}{c|}{PHPE} &\multicolumn{1}{c|}{SCA}& \multicolumn{1}{c|}{DCA-G} & \multicolumn{1}{c|}{DCA-L} & DCA-S  \\ \hline
    \multicolumn{1}{|c|}{\multirow{2}{*}{Hourglass}} & mAP & \multicolumn{1}{c|}{0.58} & \multicolumn{1}{c|}{0.50} &  \multicolumn{1}{c|}{0.36} &  \multicolumn{1}{c|}{0.18} & \multicolumn{1}{c|}{0.26} & \multicolumn{1}{c|}{0.23} & 0.24 \\ \cline{2-9} 
    \multicolumn{1}{|c|}{}                  & ATR & \multicolumn{1}{c|}{--} & \multicolumn{1}{c|}{0.33} & \multicolumn{1}{c|}{0.48} & \multicolumn{1}{c|}{\textbf{0.74}} & \multicolumn{1}{c|}{0.55} & \multicolumn{1}{c|}{0.66} & 0.63 \\ \hline
    \end{tabular}

        }
    \end{center}
\vspace{0.25cm}
\caption{\small Black-box attack results on the MS-COCO Keypoints dataset. \textbf{To:} the leftmost column denotes the target models that adversarial examples transfer to (i.e., are tested on).  \textbf{From:} the top row denotes the models which adversarial examples are generated from.}
\label{table_blackcoco}
\end{table}

\section{Conclusion}\label{sec:conclusion}
In this work, we propose the first adversarial attack on anchor-free detectors. It is a category-wise attack that attacks important pixels of all instances of a category simultaneously.
Our attack manifests in two forms, sparse category-wise attack (SCA) and dense category-wise attack (DCA), when minimizing the $L_0$ and $L_\infty$ norm-based perturbations, respectively. For SCA, it can generate sparse-imperceptible adversarial samples.  For DCA, we further provide three variants, DCA-G, DCA-L, and DCA-S, to enable a flexible selection of a specific attacking region:  the global region, the local region, and the semantic region, respectively, to improve the attack effectiveness and efficiency. Our experiments on large-scale benchmark datasets including PascalVOC, MS-COCO, and MS-COCO Keypoints indicate that the proposed methods achieve state-of-the-art attack performance and transferability on both object detection and human pose estimation tasks.

\textbf{Limitations}. There are two limitations in our proposed methods. The first limitation is that SCA has significantly higher time complexity since its Algorithm~\ref{algorithm_sca} includes CWDF and Linear Solver procedures. Our experimental results confirm it: SCA needs a much longer time to attack an image for both object detection and human pose estimation. The second limitation is that the transferability of our proposed methods on attacking SSD is much lower than that of our methods on attacking other models. 


\textbf{Future Work}.  First, we will try to address the aforementioned limitations of our proposed methods. Second, we plan to incorporate a scheme to automatically determine the hyperparameters. Third, we will try to develop effective defenses against the proposed attacks, which is important for practical applications of anchor-free detectors. 

\appendices
\section{Algorithm of LinearSolver}\label{Sec:LinearSolver}
The LinearSolver algorithm is shown in Algorithm \ref{alg:linearsolver}. In each iteration, we project towards only one single coordinate of $\w$. If projecting $\x$ to a specific direction does not provide a solution, it will be ignored in the next iteration. More details can be found in \cite{modas2019sparsefool}. Note that the projection operator of $Q$ in Algorithm \ref{alg:linearsolver} controls the pixel values between 0 and 255.  
\begin{algorithm}[ht]
    \caption{LinearSolver}
    \label{alg:linearsolver}
    \begin{algorithmic}
        \Require
        image $\x$, normal vector $\w$, boundary point $\x^B$, projection operator $Q$
        \Ensure
       perturbated point $\x^{adv}$
        \State{\textbf{Initialize}: $\x^0 \leftarrow \x$, $i \leftarrow 0$, $\mathcal{H} = \{\}$}
        \While{$\w^T(\x^i-\x^B) \neq 0$}
            \State$\rb \leftarrow \mathbf{0}$
            \State$d \leftarrow \arg\max_{j\in \mathcal{H}}|w_j|$
             \State$r_d \leftarrow \frac{|\w^T(\x^i-\x^B)|}{|w_d|}\cdot sign(w_d)$
            \State{$\x^{(i+1)} \leftarrow Q(\x^i+\rb)$} 
            \State{$\mathcal{H} \leftarrow \mathcal{H} \cup \{d\}$}
            \State{$i \leftarrow i+1$}
        \EndWhile
        \State{\textbf{return} $\x^{adv}\leftarrow \x^i$}
    \end{algorithmic}
\end{algorithm}

\bibliographystyle{IEEEtran} 
\bibliography{main} 

\begin{IEEEbiography}
[{\includegraphics[width=1in,height=1.25in]{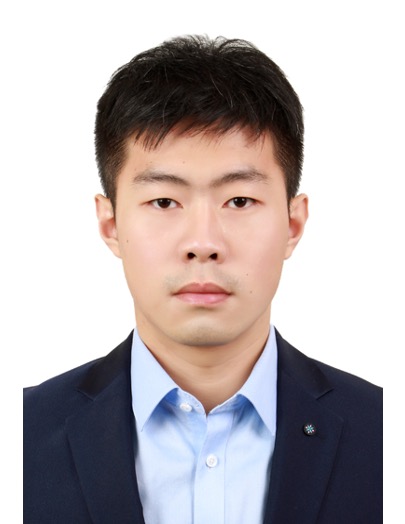}}]
{Yunxu Xie} is currently a postgraduate at the School of Computer Science, Chengdu University of  of Information Technology.
His research directions are adversarial attack, deep learning, and computer vision.
\vspace{-1.0cm}
\end{IEEEbiography}

\begin{IEEEbiography}
[{\includegraphics[width=1in,height=1.25in]{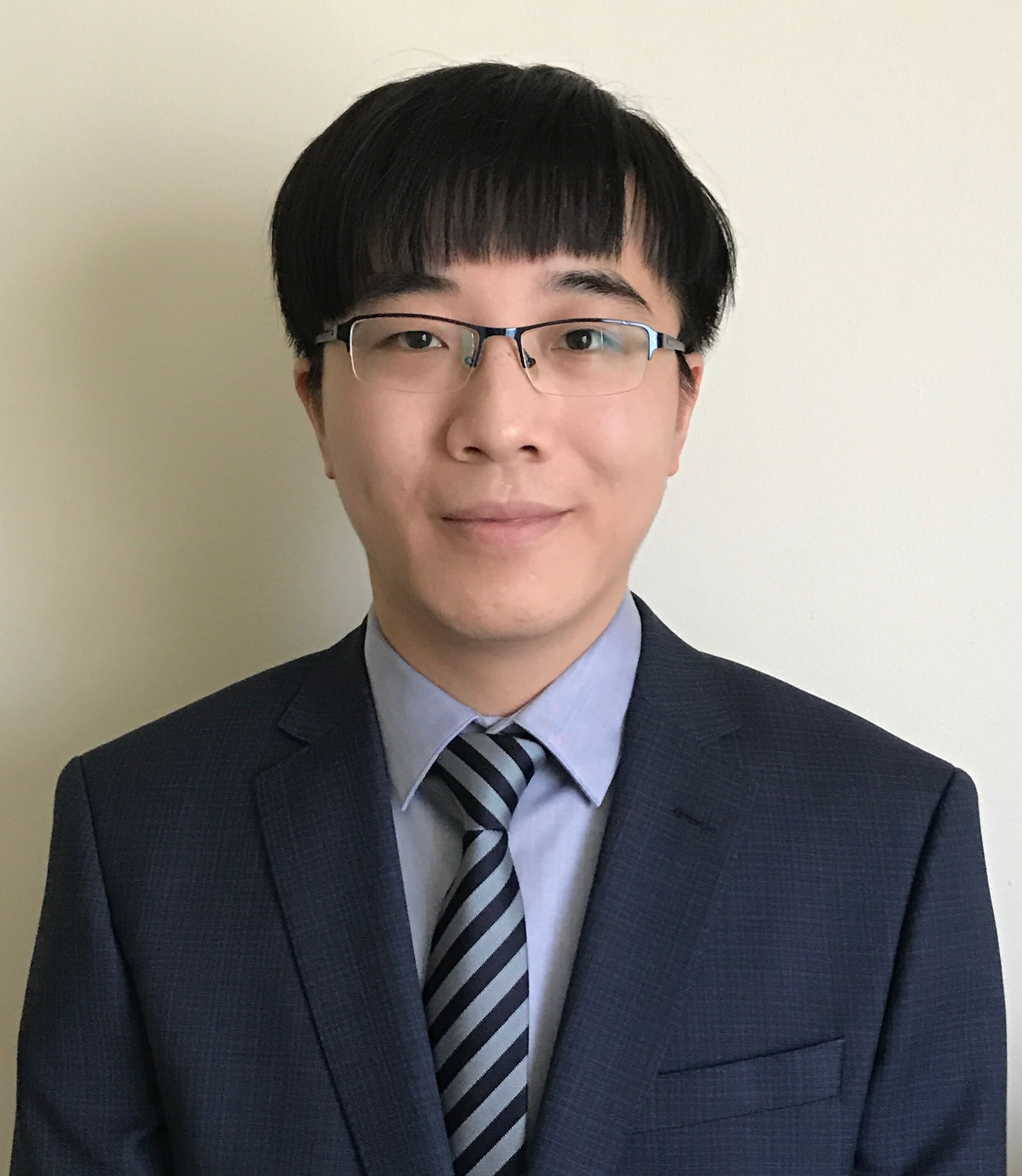}}]
{Dr. Shu Hu} is a Postdoc at Carnegie Mellon University. He received his Ph.D. degree in Computer Science and Engineering from University at Buffalo, the State University of New York (SUNY) in 2022. He received his M.A. degree in Mathematics from University at Albany, SUNY in 2020, and M.Eng. degree in Software Engineering from University of Science and Technology of China in 2016. His research interests include machine learning, digital media forensics, and computer vision. 
\vspace{-1.0cm}
\end{IEEEbiography}

\begin{IEEEbiography}
[{\includegraphics[width=1in,height=1.25in]{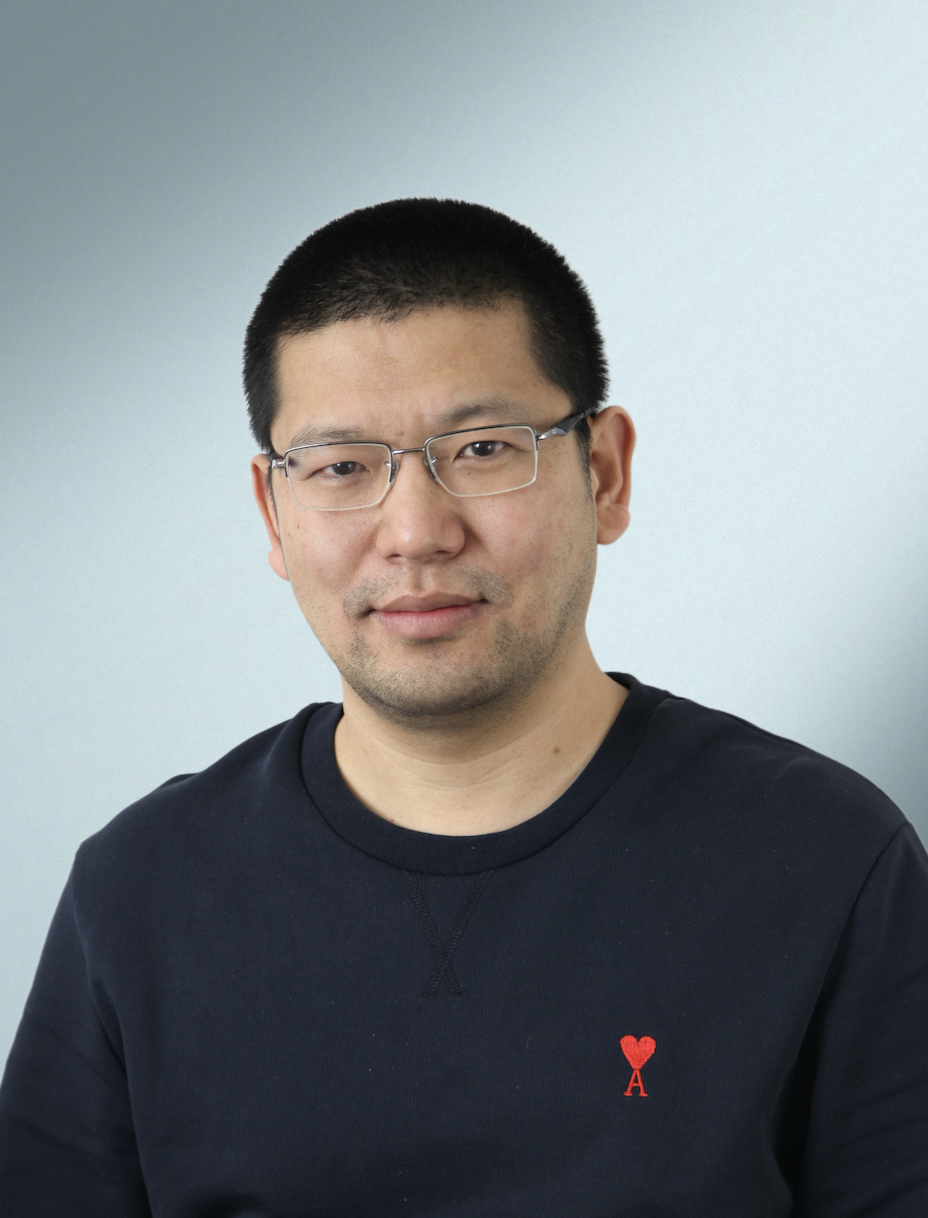}}]
{Dr. Xin Wang} (SM'2020)
is a research affiliate at University at Buffalo, State University of New York. He received his Ph.D. degree in Computer Science from the University at Albany, State University of New York in 2015. His research interests are in machine learning, reinforcement learning, deep learning, and their applications. He is a senior member of IEEE. 
\vspace{-1.0cm}
\end{IEEEbiography}

\begin{IEEEbiography}
[{\includegraphics[width=1in,height=1.25in]{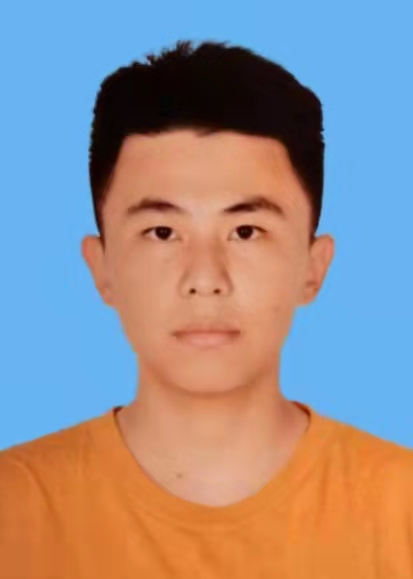}}]
{Quanyu Liao} is a graduate student at the School of Computer Science, Chengdu University of  of Information Technology.
His research directions are adversarial attack, deep learning, and computer vision.
\vspace{-1.0cm}
\end{IEEEbiography}

\begin{IEEEbiography}
[{\includegraphics[width=1in,height=1.25in]{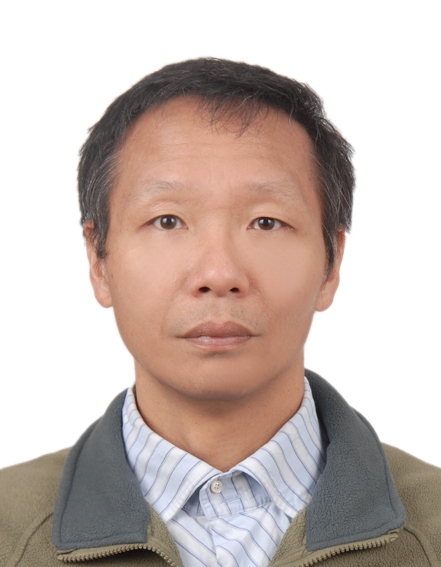}}]
{Bin B. Zhu} received the B.S. degree in physics from the University of Science and Technology of China, Hefei, China, in 1986, and the M.S. and Ph. D. degrees in electrical engineering from the University of Minnesota, Minneapolis, MN, in 1993 and 1998, respectively. He is currently a Principal Researcher with Microsoft Research Asia, Beijing, China. His research interests include DNN security and privacy, AI applications, Internet and system security, privacy-preserving processing, content protection, and signal and multimedia processing.
\vspace{-1.0cm}
\end{IEEEbiography}

\begin{IEEEbiography}
[{\includegraphics[width=1in,height=1.25in]{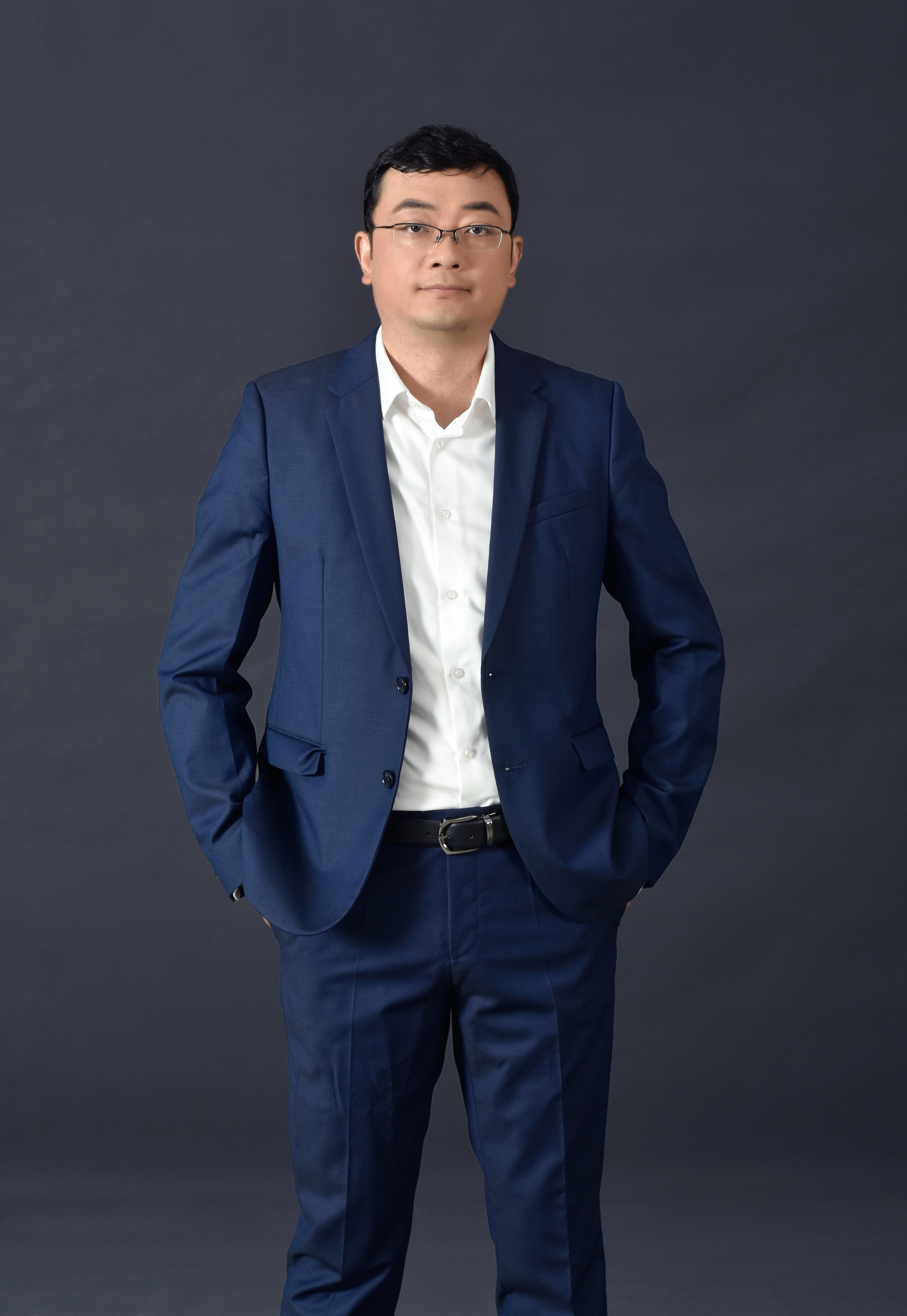}}]
{Dr. Xi Wu} is currently the dean of the School of Computer Science, Chengdu University of Information Technology, and the Chinese director of the International Joint Research Center for Image and Vision, Chengdu University of Information Technology. The main research directions are: image analysis and computational imaging, high-performance and parallel distributed computing, smart meteorology and numerical weather computing.
\vspace{-1.0cm}
\end{IEEEbiography}

\begin{IEEEbiography}
[{\includegraphics[width=1in,height=1.25in]{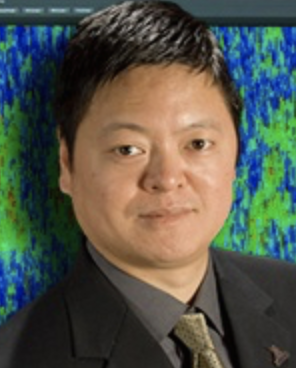}}]
{Dr. Siwei Lyu} is an SUNY Empire Innovation Professor at the Department of Computer Science and Engineering, the Director of UB Media Forensic Lab (UB MDFL), and the founding Co-Director of Center for Information Integrity (CII) of University at Buffalo, State University of New York. Dr. Lyu received his Ph.D. degree in Computer Science from Dartmouth College in 2005, and his M.S. degree in Computer Science in 2000 and B.S. degree in Information Science in 1997, both from Peking University, China.
Dr. Lyu’s research interests include digital media forensics, computer vision, and machine learning. Dr. Lyu is a Fellow of IEEE.
\end{IEEEbiography}

\end{document}